\documentclass[nohyperref]{article}

\usepackage{microtype}
\usepackage{graphicx}
\usepackage{subfigure}
\usepackage{booktabs} 
\usepackage{color}
\usepackage{kotex}
\usepackage{url}
\usepackage{bm}
\usepackage{amsfonts} 
\usepackage{array,multirow}
\usepackage{tabularx}
\usepackage{arydshln}
\usepackage{adjustbox}
\usepackage{wrapfig}
\usepackage{xcolor, pifont}
\newcommand{\cmark}{\ding{51}}%
\newcommand*\colourcheck[1]{%
  \expandafter\newcommand\csname #1check\endcsname{\textcolor{#1}{\ding{51}}}%
}
\newcommand*\colourx[1]{%
  \expandafter\newcommand\csname #1x\endcsname{\textcolor{#1}{\ding{55}}}%
}
\colourcheck{blue}
\colourcheck{green}
\colourx{gray}

\renewcommand{\arraystretch}{1.02}
\usepackage{hyperref}

\usepackage[accepted]{icml2022}

\usepackage{amsmath}
\usepackage{amssymb}
\usepackage{mathtools}
\usepackage{amsthm}

\usepackage[capitalize,noabbrev]{cleveref}

\theoremstyle{plain}
\newtheorem{theorem}{Theorem}[section]

\theoremstyle{definition}
\newtheorem{definition}[theorem]{Definition}

\theoremstyle{remark}

\usepackage{algorithmic}

\definecolor{customred}{RGB}{192, 0, 0}

\newcommand*\Let[2]{\STATE #1 $\gets$ #2}

\usepackage[textsize=tiny]{todonotes}

\icmltitlerunning{TAM: Topology-Aware Margin Loss for Class-Imbalanced Node Classification}

\begin{document}

\twocolumn[
\icmltitle{TAM: Topology-Aware Margin Loss for Class-Imbalanced Node Classification}



\icmlsetsymbol{equal}{*}

\begin{icmlauthorlist}
\icmlauthor{Jaeyun Song}{equal,KAIST}
\icmlauthor{Joonhyung Park}{equal,KAIST}
\icmlauthor{Eunho Yang}{KAIST,aitrics}
\end{icmlauthorlist}

\icmlaffiliation{KAIST}{Graduate School of AI, Korea Advanced Institute of Science and Technology (KAIST), Daejeon, South Korea}
\icmlaffiliation{aitrics}{AITRICS, Seoul, South Korea}

\icmlcorrespondingauthor{Jaeyun Song}{mercery@kaist.ac.kr}
\icmlcorrespondingauthor{Joonhyung Park}{deepjoon@kaist.ac.kr}
\icmlcorrespondingauthor{Eunho Yang}{eunhoy@kaist.ac.kr}

\icmlkeywords{Machine Learning, ICML}

\vskip 0.3in
]



\printAffiliationsAndNotice{\icmlEqualContribution} 

\begin{abstract}
Learning unbiased node representations under class-imbalanced graph data is challenging due to interactions between adjacent nodes. Existing studies have in common that they compensate the minor class nodes `as a group' according to their overall quantity (ignoring node connections in graph), which inevitably increase the false positive cases for major nodes. We hypothesize that the increase in these false positive cases is highly affected by the label distribution around each node and confirm it experimentally. In addition, in order to handle this issue, we propose Topology-Aware Margin (TAM) to reflect local topology on the learning objective. Our method compares the connectivity pattern of each node with the class-averaged counter-part and adaptively adjusts the margin accordingly based on that. Our method consistently exhibits superiority over the baselines on various node classification benchmark datasets with representative GNN architectures.

\end{abstract}

\section{Introduction}
The importance of learning qualitative node representation has been emerging to accurately classify the node property in real-world graphs such as social networks, commercial graphs, and chemical molecules~\citep{DBLP:journals/scn/MohammadrezaeiS18, DBLP:conf/kdd/YingHCEHL18, sage}. Recently, graph neural networks (GNNs)~\citep{gcn, gat, sage} are widely adopted to handle graph-structured data and have shown remarkable success in various fields. However, as natural graphs could be class-imbalanced inherently, GNNs are prone to be biased toward major classes.
Learning from those graphs without handling class-imbalanced issue leads to low accuracy for minor classes. Although the simple solution is to curate class-balanced graphs, collecting data in a balanced way is not always possible. 



To address this problem, diverse imbalance handling strategies for node classification ~\citep{conditionalgan, graphsmote, imgagn, graphens} have been recently proposed. These methods fortify minor classes in their own way such as
extending SMOTE method~\citep{smote} to graph-structured data~\citep{graphsmote}, mixing nodes by considering neighbor structure~\citep{graphens} or generating virtual minor nodes via the conditional GAN~\citep{conditionalgan}. 

However, these approaches overlook the fact that when compensating the minor classes based on their quantity, certain nodes could significantly degrade the performance of other classes. Considering the innate characteristics of message passing algorithms of GNNs, we hypothesize that the entire representation learning procedure can be misled by weighted minor nodes in the aggregation of message passing and that the effect is more attributed to nodes with high connectivity rates with other (major) classes. Toward this direction, we observe that compensating such minor nodes with high connectivity rates to major class significantly increase false positives for major nodes. In line with this observation, we confirm that existing imbalance handling algorithms show sub-optimal performances as they do not reflect this local topology when weighting the minor classes. 
Although not directly related to our hypothesis, ReNode~\citep{renode} is somehow related in terms of adjusting the weights of some nodes; this method decreases the weights of nodes close to topological class boundaries. However, this method can only work for homogeneously-connected graphs. Moreover, since the connectivity patterns and class-wise relative weighting for multi-class cases are not considered in the weighting process, the impacts of individual nodes affecting other classes are still not properly identified.

Armed with this hypothesis, in this paper we propose \textbf{T}opology-\textbf{A}ware \textbf{M}argin (TAM), a node-wise logit adjustment method, which takes into account their local topology in terms of class-pair connectivity and neighbor distribution statistics. 
Our key principle is as follows: if a (minor) node is highly likely to be confused with specific (major) classes considering its local topology, we should decrease the margins for those (major) classes so that GNNs can be trained in a well-calibrated manner (informally speaking, when some minor node has abnormally many major neighbors, we reduce the weight for it). Toward this, first we devise Anomalous Connectivity-aware Margin (ACM) that decreases the target class margin of a node if it has relatively high neighbor density for that target class. At the same time, we introduce Anomalous Distribution-aware Margin (ADM) that calculates the degree of confusion based on the average neighbor statistics of \emph{target} class and additionally adjusts the margin of the target class.

Our method can be combined with most imbalance handling approaches seamlessly and consistently brings the performance enhancement over multiple node classification benchmark datasets such as citation networks~\citep{DBLP:journals/aim/SenNBGGE08}, WebKB, and Wikipedia networks~\citep{wikipedia} with various architectures including GCN~\citep{gcn}, GAT~\citep{gat}, and GraphSAGE~\citep{sage}.

Our contribution is threefold:
\begin{itemize}
    \item We hypothesize and confirm that false positives due to compensating minor nodes do not appear evenly on the graph, and are highly affected by the neighbor label distribution around each node. Specifically, we demonstrate that a significantly high false positive ratio appears around minor nodes that have higher connectivity with major nodes. 
    \item We propose a tailored solution to this hypothesis that can effectively decrease excessive false positives by individually adjusting the extent of compensation based on node topology compared to class statistics.
    \item Our method can be combined with existing imbalance handling methods regardless of their compensating strategies. When combined with our method, baselines consistently improve the imbalance handling performance on multiple benchmark datasets. 
\end{itemize}

\section{Preliminary}

\subsection{Notation and Definitions}
We target a semi-supervised node classification task on an undirected graph $G(V,E)$ where $V$ is a node set, $E$ is the set of edges, and $\mathcal{Y}$ is the set of possible class labels. $Y$ is set of labels for $V$ and $V^L$ is the set of labeled nodes ($V^L\subseteq V$). $X\in \mathbb{R}^{\vert V\vert\times d}$ is the node feature matrix where the $i$-th node $v_i$ has the node feature $x_i$ (the $i$-th row of $X$). Let $\mathcal{N}(v)$ be the set of adjacent nodes to node $v$: $\{u\in V|{u,v}\in E\}$. $d_v$ is the degree of node $v$: $\vert\mathcal{N}(v)\vert$.

We here introduce two key definitions leveraged for estimating the node- and class-level connectivity: neighbor label distribution ($\mathcal{D}$) and class-wise connectivity matrix ($\mathcal{C}$).
\begin{definition}
    \textbf{(Neighbor Label Distribution $\mathcal{D}$).} Let $\mathcal{D}\in \mathbb{R}^{\vert V\vert\times\vert \mathcal{Y}\vert}$. Then, neighbor label distribution (NLD) $\mathcal{D}$ is defined as:
    \begin{align}
        \mathcal{D}_{i,j} = \frac{\vert\{v\in \mathcal{N}(i)\cup\{i\}|y_v=j\}\vert}{d_i+1}. \label{def:nld}
    \end{align}
\end{definition}
That is, the $i$-th row represents the distribution of neighbor labels of node $i$ (including node $y_i$ itself).

\begin{definition}
    \textbf{(Class-wise Connectivity Matrix $\mathcal{C}$).} Let $\mathcal{C}\in \mathbb{R}^{\vert \mathcal{Y}\vert\times\vert \mathcal{Y}\vert}$. Then, class-wise connectivity matrix $\mathcal{C}$ is defined as:
    \begin{align}
        \mathcal{C}_{i,j} = \frac{1}{\vert\{v\in V|y_v=i\}\vert} \sum_{u\in\{v\in V|y_v=i\}} \mathcal{D}_{u,j}. \label{def:ccm}
    \end{align}
\end{definition}
We define these two terms similarly with homophily used in \citet{geomgcn} since our method requires to compute neighbor label distribution for a given node and its class statistics. Note that two concepts are computed under the assumption that neighbor label distributions for all nodes are given. However, this assumption is hardly satisfied in real world, hence we utilize model prediction for unlabeled nodes in main experiments (Section~\ref{subsec:cls-temperature}).

\subsection{Node Classification with Graph Neural Networks}

In this section, we briefly describe the GNNs in terms of node classification. The $l$-th layer of GNNs consists of three functions including message function $m_l$, feature aggregation function $\psi_l$, and node feature update functions $\gamma_l$. For node $v$, node feature $x^{(l+1)}_v$ is derived from  $x^{(l)}_v$ as follows.
\begin{equation}
    \resizebox{0.95\linewidth}{!}{$
    x^{(l+1)}_v=\gamma_l\left(x^{(l)}_v,\psi_l\left(\{m_l(x^{(l)}_v,x^{(l)}_u,w_{v,u})|u\in\mathcal{N}(v)\}\right)\right)
    $}
\end{equation}
where $w_{u,v}$ is the edge weight of edge $\{v,u\}\in E$. For example, node feature $x^{(l)}_v$ of Graph Convolutional Network (GCN)~\citep{gcn} is computed as $x^{(l+1)}_v=\sum_{u\in\mathcal{N}(v)\cup\{v\}}\frac{\Theta_l^\top x^{(l)}_u}{\sqrt{\hat{d_v}\hat{d_u}}}$, where $\hat{d}_v =  1 + \sum_{u \in \mathcal{N}(v)} {w_{v, u}}$ and $\Theta_l$ is the matrix of filter parameters at the $l$-th layer.

\begin{figure*}[t]
  \centering
   \begin{minipage}[b]{0.246\textwidth}
    \includegraphics[width=\textwidth]{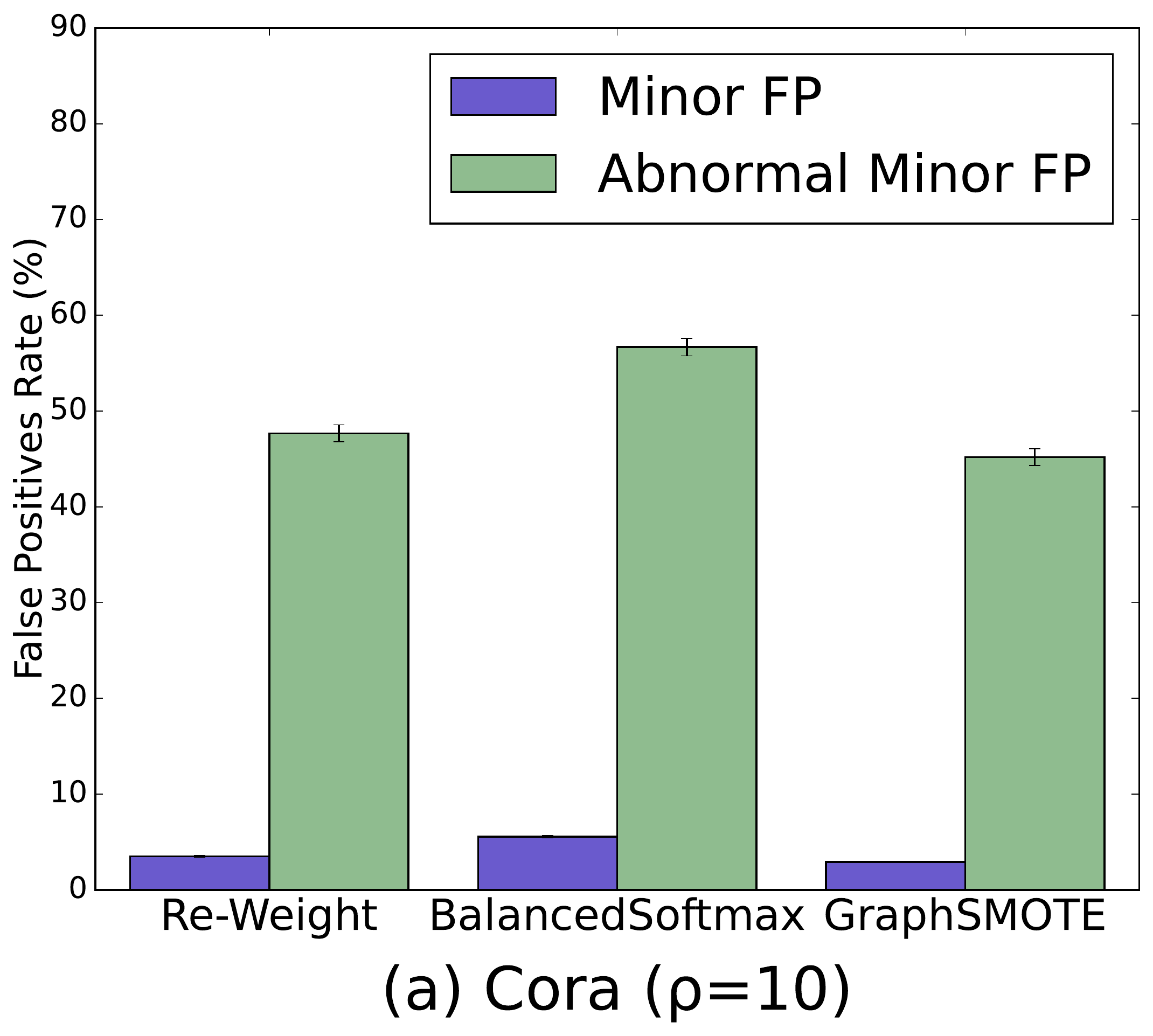}
  \end{minipage}
  \begin{minipage}[b]{0.246\textwidth}
    \includegraphics[width=\textwidth]{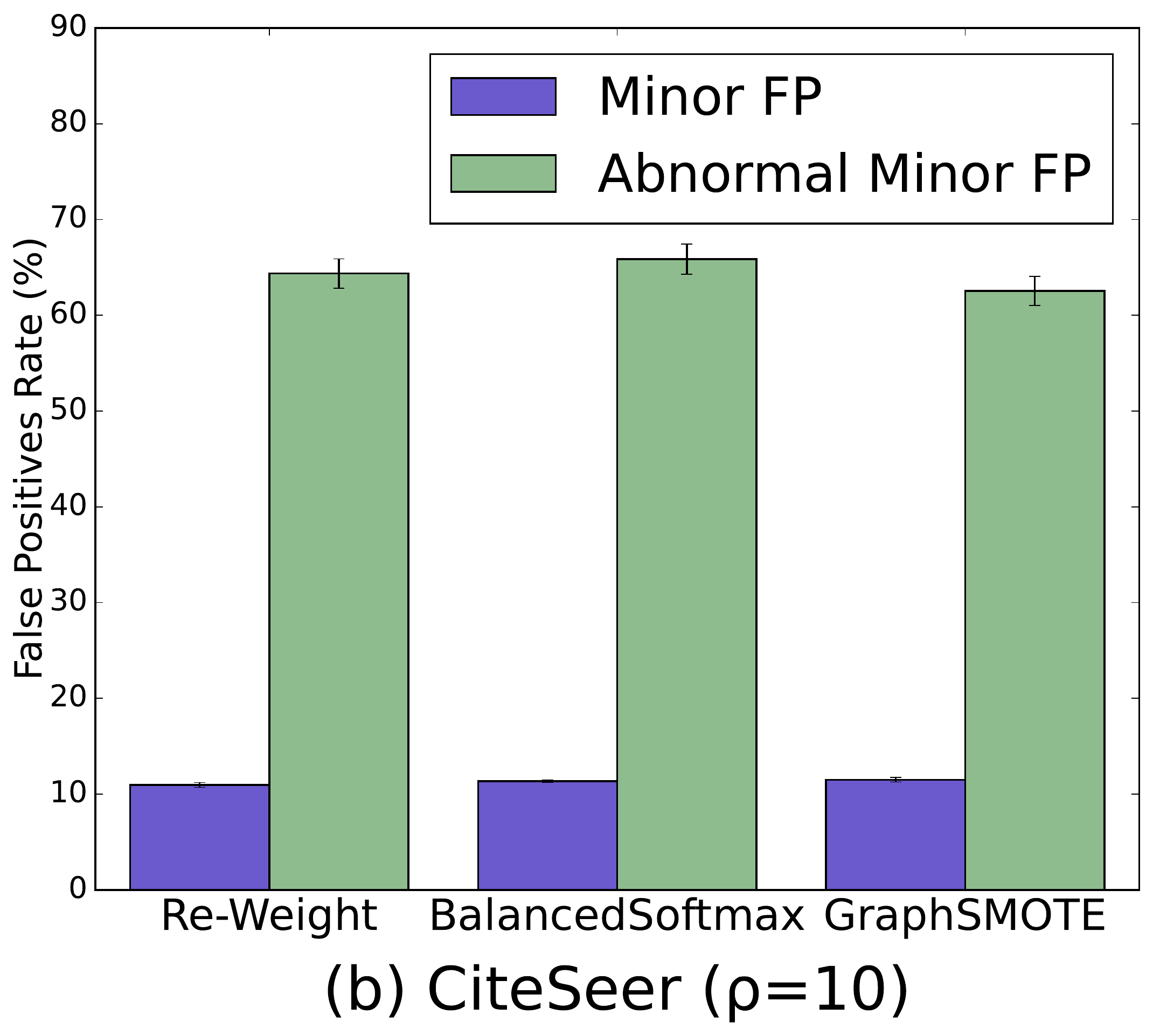}
  \end{minipage}
  \begin{minipage}[b]{0.246\textwidth}
    \includegraphics[width=\textwidth]{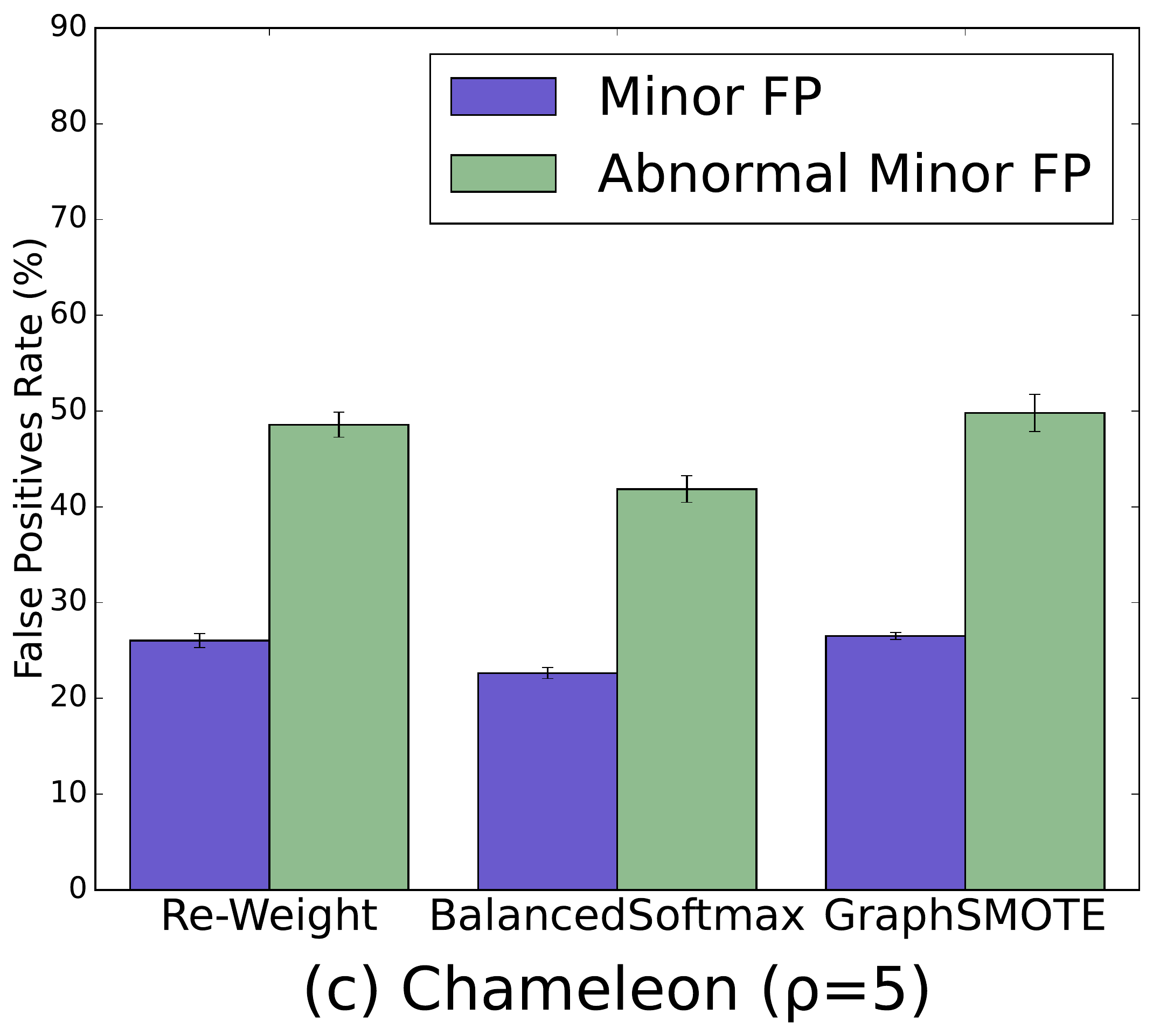}
  \end{minipage}
  \begin{minipage}[b]{0.246\textwidth}
    \includegraphics[width=\textwidth]{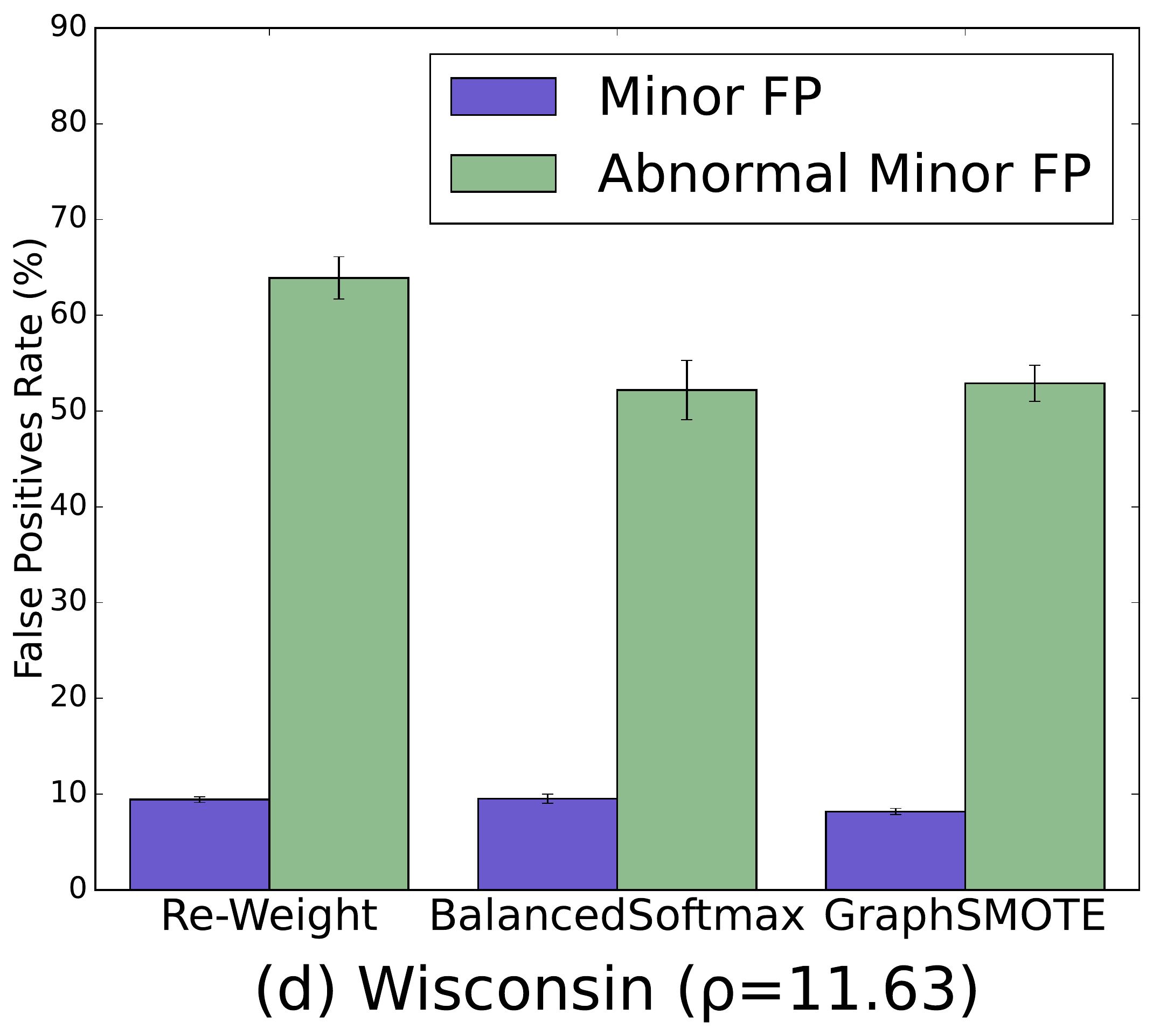}
  \end{minipage}
  
  \begin{minipage}[b]{0.245\textwidth}
    \includegraphics[width=\textwidth]{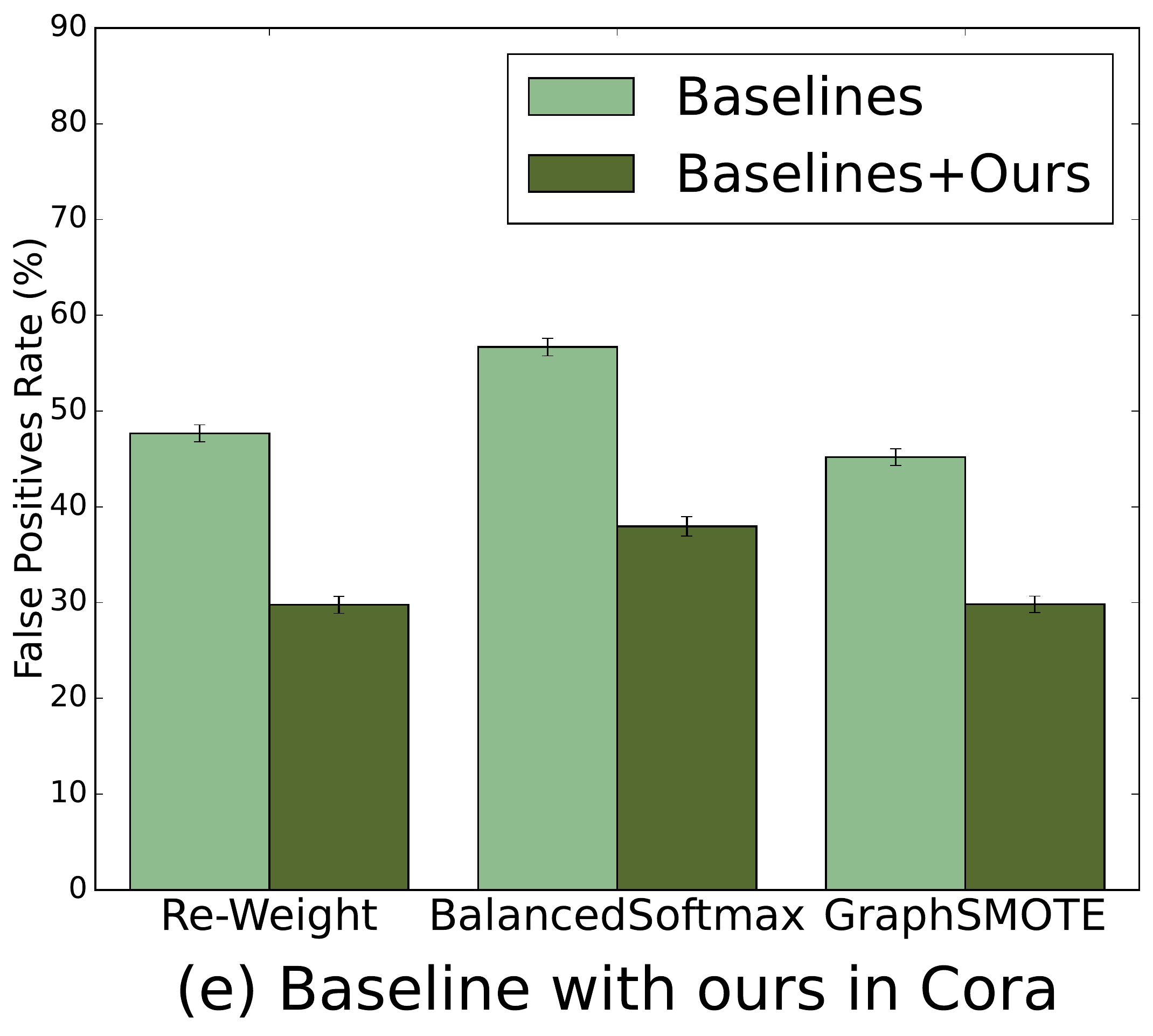}
  \end{minipage}
  \begin{minipage}[b]{0.245\textwidth}
    \includegraphics[width=\textwidth]{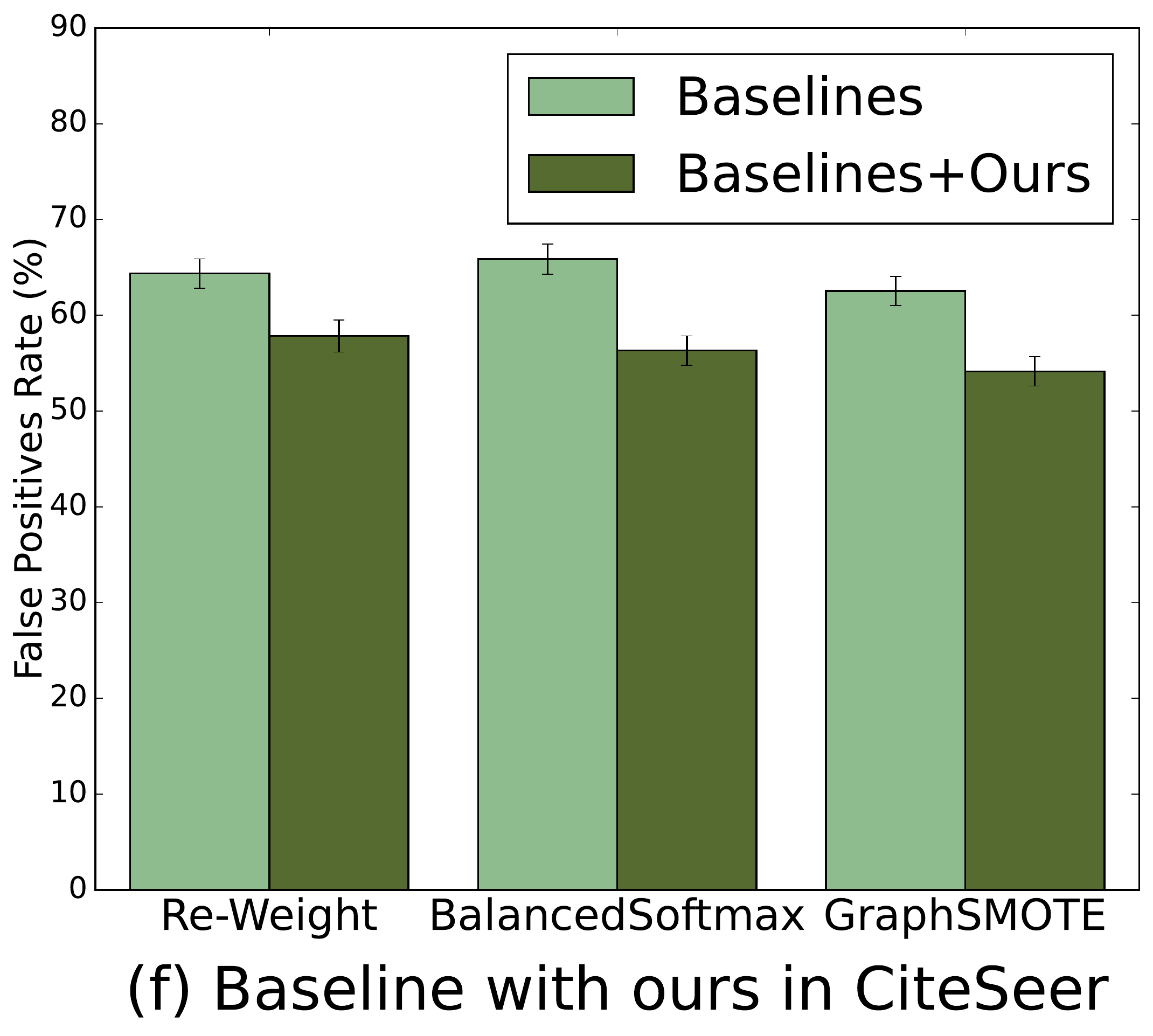}
  \end{minipage}
  \begin{minipage}[b]{0.249\textwidth}
    \includegraphics[width=\textwidth]{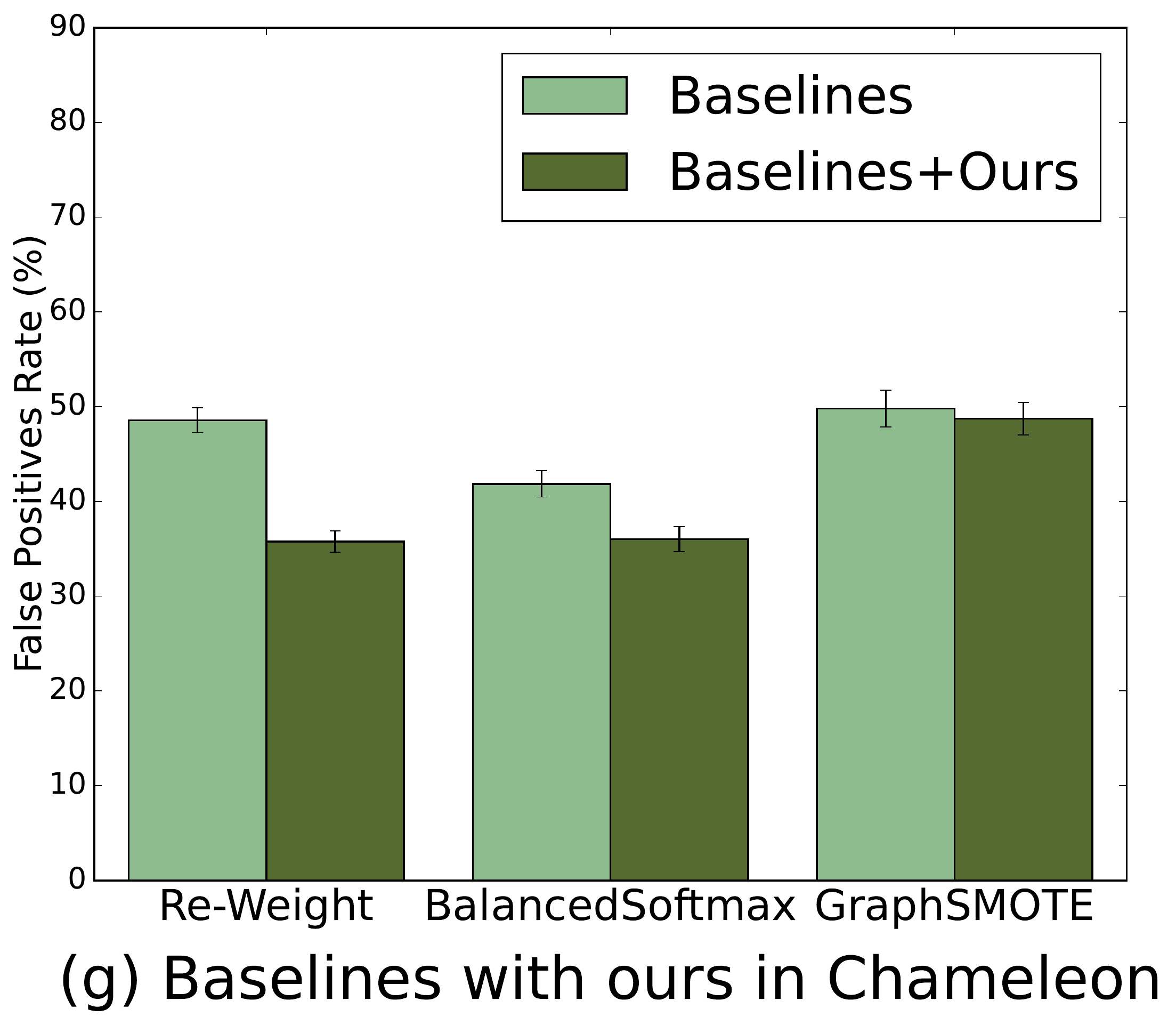}
  \end{minipage}
  \begin{minipage}[b]{0.245\textwidth}
    \includegraphics[width=\textwidth]{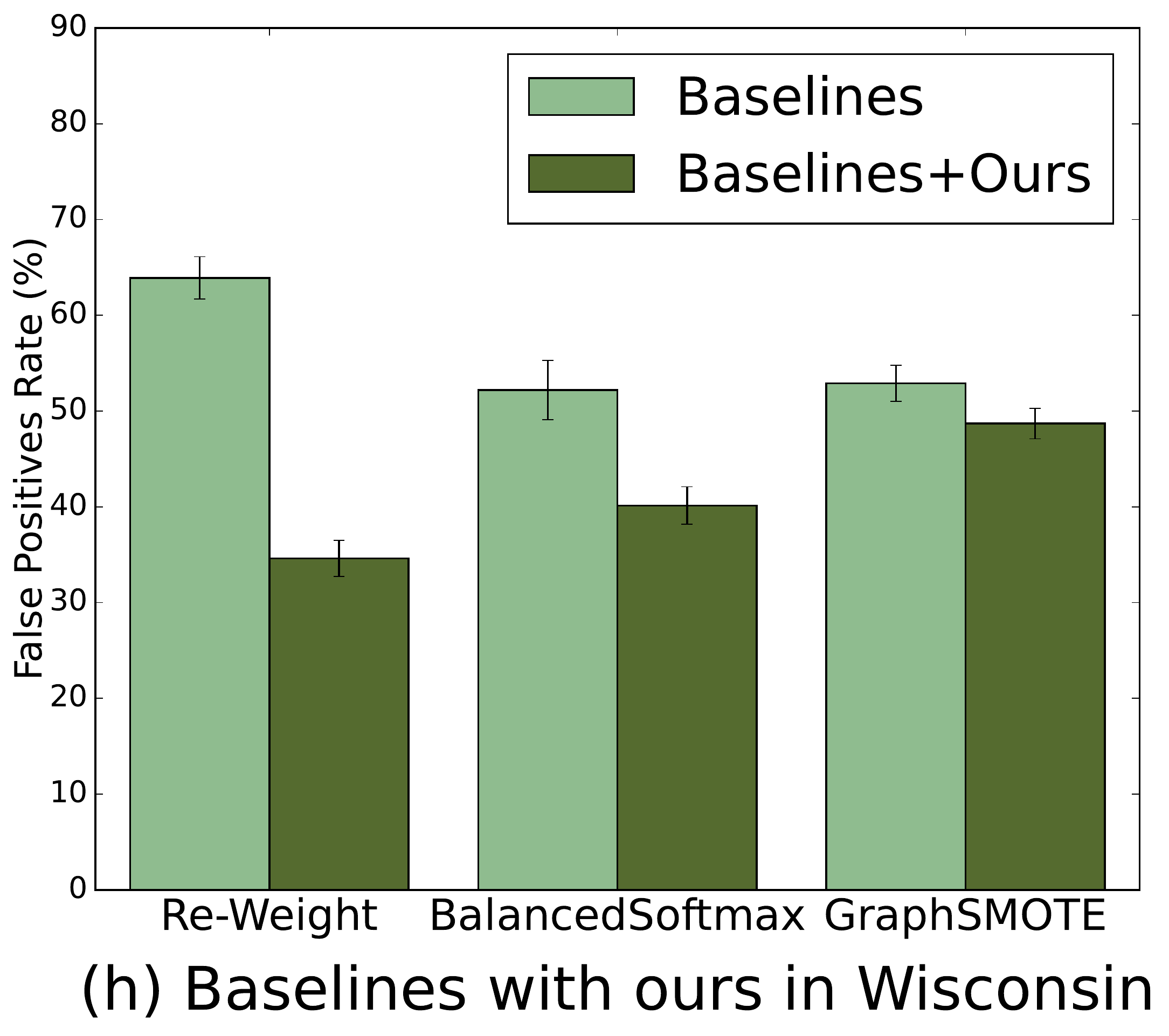}
  \end{minipage}
  
  \vspace{-0.15in}
  \caption{\small Comparison of false positive rates near normal minor nodes and anomalously-connected minor nodes. For (a) $\sim$ (d), \textbf{Abnormal Minor FP} represents false positive rates when major nodes are connected with anomalous minor nodes. \textbf{Minor FP} presents the average probability of being false positives. In (e) $\sim$ (h), the results of the change in false positive rates (caused by abnormal minor nodes) are presented when baselines are integrated with our method. }
  \vspace{-0.05in}
  \label{fig:fp_topology}
\end{figure*}
\subsection{Margin-based Class-Imbalance Handling}
We revisit margin-based imbalance handling methods in the vision domain~\citep{ldam, equalization, balanced_softmax, logit_adjustment}. Margin-based approaches alleviate the bias to major classes by increasing the margin of minor classes to major classes or decreasing the margin of major classes to minor classes in the training phase and show significantly superior performance than other loss modification algorithms~\citep{balanced_softmax}. Specifically, let us define the quantity of the $k$-th class $N_k$, then cross entropy (CE) with Balanced Softmax~\citep{balanced_softmax} is computed for node $v$ as:
\begin{align}
    \mathcal{L}=\mathcal{L}_{CE}(l_v+m,y_v)=-\text{log}\left(\frac{e^{l_{v,y_v}+\text{log}N_{y_v}}}{\sum_{k\in\mathcal{Y}}e^{l_{v,k}+\text{log}N_k}}\right), \label{eq:bs}
\end{align}
where $l_v, y_v$ are the logit and the label of node $v$ respectively, and $m=(\text{log}N_1,\text{log}N_2,\cdots,\text{log}N_{\vert\mathcal{Y}\vert})$. In multi-class Softmax regression, Balanced Softmax minimizes the generalization bound~\citep{balanced_softmax}. In that margin-based approaches could adjust logits by considering the relative quantity ratio between two classes and are effective in the vision domain, we adopt margin-based approaches in our algorithm.

\section{Analysis of Anomalous Connectivity}~\label{sec:problem}
Our primary research hypothesis is minor nodes that deviate from the connectivity pattern induce excessive false positives during the quantity-based compensating process. To verify our assumption empirically, we investigate the \textit{topological positions of false positives} on minor classes.
\paragraph{Experimental Design} 
We design an experiment to compare the false positives ratios on neighbors of anomalously connected minor nodes with those of normal minor nodes assuming that neighbor label information is accessible. First, we define anomalously connected node set $V^{*}$ as $V^{*} = \{v\in V^L \vert \underset{\scriptscriptstyle{t\in  \vert \mathcal{Y}\vert \setminus \{y_v\}}}{max}\frac{\mathcal{D}_{v,t}}{\mathcal{C}_{y_v,t}} > 1   \}$, which is a set of nodes that has more edge connections with other classes compared to class-averaged level. $V^{*}_{minor}\subset V^{*}$ is a set of minor class nodes belonging to $V^{*}$, $V_{major} \subset {V \setminus V^L}$ is a set of major nodes in validation set, and $FP(\cdot)$ is a function that counts the number of false positives for the minor classes.

We calculate the ratio $\frac{FP(\{\mathcal{N}(v) \cap V_{major} \vert v \in V_{minor}^* \})}{\vert \mathcal{N}(v) \cap V_{major} \vert v \in V_{minor}^* \vert }$, representing the probability of being false positives when major nodes are connected with anomalous minor nodes (\textbf{\footnotesize Abnormal Minor FP} in Figure~\ref{fig:fp_topology}). Then we compare the computed probability with the ratio $\frac{FP(V_{major})}{\vert V_{major} \vert}$, the average probability of being false positives (\textbf{\footnotesize Minor FP} in Figure~\ref{fig:fp_topology}).
Experimental details are described in the following paragraph.

\paragraph{Settings}
We conduct experiments on two well-known node classification benchmark datasets - CiteSeer (homophilous graph) and Wisconsin (heterophilous graph) using GCN architecture. For CiteSeer dataset~\citep{DBLP:journals/aim/SenNBGGE08}, we follow the split of ~\citet{fastgcn} and process the label distribution to follow step imbalance setting as existing works~\citep{ldam,renode}. In other words, all minor classes have $n_{minor}$ labeled nodes and the major class nodes have $\rho * n_{minor}$ where $\rho$ is an imbalance ratio. We set $\rho$ to 10. All experiments are repeated 100 times.


\begin{figure*}[t]
\centering
\includegraphics[width=1.\linewidth]{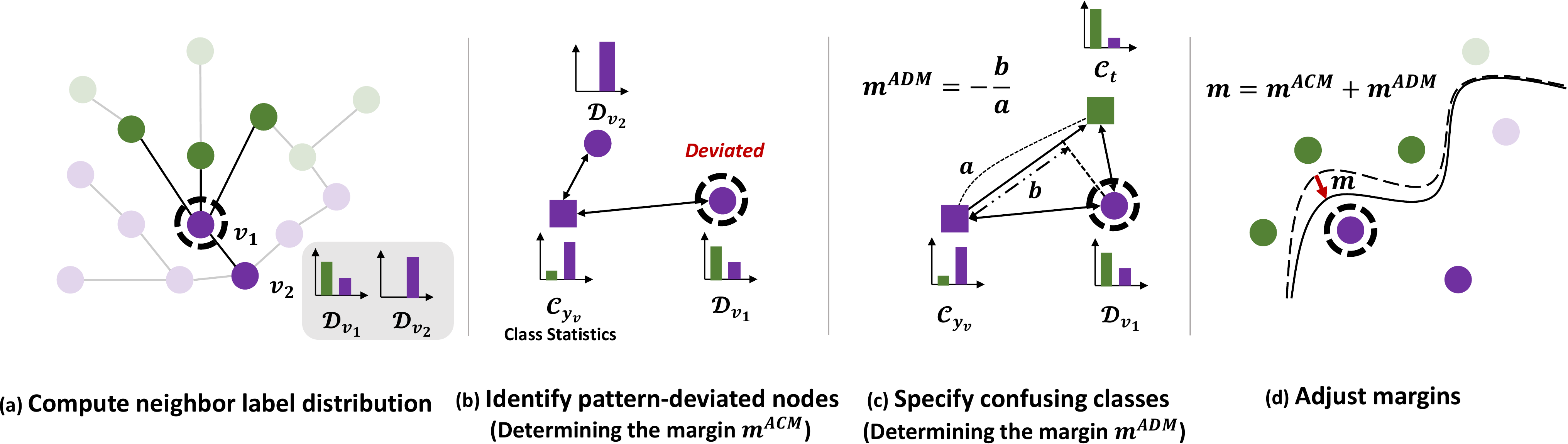}
\vspace{-0.18in}
\caption{\small The overall pipeline of TAM. We first calculate neighbor label distribution $\mathcal{D}$ by utilizing model prediction for unlabeled neighbors, then compute class-wise connectivity matrix $\mathcal{C}$ in (a). According to $\mathcal{D}$ and $\mathcal{C}$, determine ACM $m^{ACM}$ in (b) and ADM $m^{ADM}$ in (c). By applying two margins to logits, we adjust margins in (d).}
\vspace{-0.15in}
\label{fig:concept_figure}
\end{figure*}

\paragraph{Results}
To verify our assumption, we scrutinize three representative imbalance handling approaches: Re-Weight~\cite{reweight}, Balanced Softmax~\cite{balanced_softmax}, and GraphSMOTE~\cite{graphsmote}. In Figure~\ref{fig:fp_topology} (a) and (b), we confirm that false positives on minor classes are intensively concentrated around minor nodes that have higher connectivity with other classes (compared to class-averaged level)  regardless of each baseline's compensating strategy. Interestingly, the aptness of false positives is consistently exhibited in both homogeneously- and heterogeneously-connected graphs. It is worth noting that the increase in false negative rates is negligible to the decrease in false positive rates in Appendix~\ref{appensub:problem}.


\section{Proposed Method}~\label{sec:method}
We now introduce our effective margin adjustment strategy, TAM, which determines the intensity of imbalance compensation based on the local topology of individual node. In Section~\ref{sec:problem}, we investigated that reinforcing minor nodes connected more with other classes than class-averaged level induces the false positives of minor classes. Inspired by this observation, we identify topologically improbable nodes and adaptively adjust the margins for those nodes. We devise two core components of TAM. First, Anomalous Connectivity Margin (ACM) decreases the class margin of a target node (one of the neighbor nodes) if the portion of the class of target node in neighbor label distribution (NLD) is larger than class-averaged connectivity (Section~\ref{subsec:acm}). Then, Anomalous Distribution-aware Margin (ADM) adjusts the margin according to the relative distance computed using both target class-averaged NLD and self class-averaged NLD (Section~\ref{subsec:adm}). The overall pipeline and full algorithm of TAM are provided in Figure~\ref{fig:concept_figure} and in Algorithm~\ref{alg:TAM}, respectively.

Our learning objective function is formulated as:
\begin{align}
    \mathcal{L}_{TAM}=\frac{1}{\vert V^{L} \vert}\sum_{v \in V^{L}}\mathcal{L}\left(l_{v} + \alpha m^{ACM}_{v} + \beta m^{ADM}_{v}, y_{v}\right), \label{eq:loss}
\end{align}
where $\mathcal{L}$ is the loss function (such as cross entropy), $l_{v} \in \mathbb{R}^{\vert\mathcal{Y}\vert}$ is the logit of node $v$, and $y_v$ is the label of node $v$. 
Our novel components add the margins here: $m^{ACM}_{v} \in \mathbb{R}^{\vert\mathcal{Y}\vert}$ represents the ACM term of node $v$ and $m^{ADM}_{v}\in \mathbb{R}^{\vert\mathcal{Y}\vert} $ is the ADM term, with respective hyperparameters $\alpha$ and $\beta$. 

It is worth noting that our method can be orthogonally combined with any imbalance handling methods and GNNs since our method simply adjusts output logits of the model before evaluating the loss function. We describe how to compute $m^{ACM}_{v}$ and $m^{ADM}_{v}$ in the following subsections. 

\subsection{Anomalous Connectivity-Aware Margin}~\label{subsec:acm}
To restrain the generation of false positives caused by abnormally connected nodes, we suggest ACM, which modifies the margin of each class by calibrating the deviation of $\mathcal{D}_{v,:}$ (NLD of node $v$) from $\mathcal{C}_{y_v,:}$ (connectivity pattern of class $y_v$). Given a node $v$, we first compare $\mathcal{D}_{y_v,y_v}$ with averaged homophily ratio of class $y_v$, $\mathcal{C}_{y_v,y_v}$, to estimate how much node $v$ follows the class-homophily. If $\frac{\mathcal{C}_{y_v,y_v}}{\mathcal{D}_{v,y_v}}$ is high, we decreases the margins on entire classes. The intuition behind here is: as nodes that do not follow the class-homophily tendency would be risky in the imbalance handling process, we make learning signals of these nodes weak in the training phase.

To further control the margin for each class $t$, we calculate the connecting ratio with class $t$ over class-averaged level: $\frac{\mathcal{D}_{v,t}}{\mathcal{C}_{y_v,t}}$ . The high $\frac{\mathcal{D}_{v,t}}{\mathcal{C}_{y_v,t}}$ implies that node $v$ has a fair chance to be confused with class $t$ considering the message-passing of GNNs. Hence, we decreases the margin of class $t$ in this case so that GNNs can be trained in a well-calibrated manner. 

Based on this motivation, the ACM of node $v$ on class $t$ is derived as:
\begin{align}
    m^{ACM}_{v,t}= -\text{max}\left(\text{log}\left((
    \frac{\mathcal{C}_{y_v,y_v}}{\mathcal{D}_{v,y_v}}) \cdot
    (\frac{\mathcal{D}_{v,t}}{\mathcal{C}_{y_v,t}})
    \right),0\right) \label{eq:acm}
\end{align}
Note that the margin of its own class $m^{ACM}_{v,y_v}$ is set to 0 by the above equation. 

\begin{algorithm}[t]
    \footnotesize
	\caption{Topology-Aware Margin}
	\label{alg:TAM}
	{
	\begin{algorithmic}[1]
			\STATE {\bfseries Input:} Graph $G(X,V,E,Y)$, set of possible class labels $\mathcal{Y}$, set of labeled nodes $V^L$, model $f_\theta$, loss function $\mathcal{L}$, hyperparameters $\alpha,\beta$, learning rate $\eta$, label $y_v$ onehot vector, $e_{y_v}$.
			\STATE {\bfseries Initialize:} Model parameter $\theta \in \mathbb{R}^{d}$.
			\STATE {Compute $T_1,T_2,\cdots,T_k$ following the Equation~\eqref{eq:classtemp}}
			\FOR {$t = 1, 2, \ldots, T$}
    			\Let{$l$}{$f_\theta(X,V,E)$} \COMMENT{Model prediction}
    			\FOR {$k = 1, 2, \cdots, \vert\mathcal{Y}\vert$} 
                    \Let{$p_{:,k}$}{$\frac{l_{:,k}}{T_k}$} 
    			\ENDFOR \COMMENT{Class-wise temperature}

    			\Let{$\mathcal{D}$}{\textbf{0}}
    			\FOR {$v\in V^L$}
    			    \FOR {$u\in\mathcal{N}(v)\cup\{v\}$}
        			    \IF {$u\in V^L$}
            			    \Let{$\mathcal{D}_{v,:}$}{$\mathcal{D}_{v,:}+\frac{1}{\vert\mathcal{N}(v)\vert+1}e_{y_u}$}
            			\ELSE
            			   \Let{$\mathcal{D}_{v,:}$}{$\mathcal{D}_{v,:}+\frac{1}{\vert\mathcal{N}(v)\vert+1}\text{Softmax}(p_{u,:})$}
            	        \ENDIF
        	        \ENDFOR
    			\ENDFOR \COMMENT{Neighbor label distribution}

    			\FOR {$k = 1, 2, \cdots, \vert\mathcal{Y}\vert$}
    			    \Let{$\mathcal{C}_{k,:}$}{$\frac{1}{\vert\{v\in V^L|y_v=k\}\vert} \sum_{u\in\{v\in V^L|y_v=k\}} \mathcal{D}_{u,:}$}
    			\ENDFOR \COMMENT{Class-wise connectivity matrix}

    			\FOR {$v\in V^L$}
    			    \FOR {$k = 1, 2, \cdots, \vert\mathcal{Y}\vert$}
    			        \STATE {Compute $\cos A_{v,k}$ following the Equation~\eqref{eq:cos}}
    			        \Let{$m^{ACM}_{v,k}$}{$-\text{max}\left(\text{log}\left((\frac{\mathcal{C}_{y_v,y_v}}{\mathcal{D}_{v,y_v}}) \cdot (\frac{\mathcal{D}_{v,t}}{\mathcal{C}_{y_v,t}}) \right),0\right)$}
    			        \Let{$m^{ADM}_{v,k}$}{$-\frac{\text{JS}\left(\mathcal{D}_{v,:},\mathcal{C}_{y_v,:}\right)\cos A_{v,k}} {\text{JS} \left(\mathcal{C}_{t,:}, \mathcal{C}_{y_v,:} \right)}$}
    			    \ENDFOR
    			\ENDFOR \COMMENT{ACM \& ADM}

    			\Let{$\mathcal{L}_{TAM}$}{$\mathcal{L}(l^L+\alpha m^{ACM}+\beta m^{ADM},Y^L)$}
    			\Let{$\theta$}{$\theta - \eta \nabla\mathcal{L}_{TAM}$}
			\ENDFOR
			\STATE {\bfseries Output:} $\theta$
	\end{algorithmic}
	}
\end{algorithm}

\subsection{Anomalous Distribution-Aware Margin}~\label{subsec:adm}
Although ACM can identify nodes that deviated from connectivity patterns, it is not sufficient to recognize whether a deviated node is confused with other classes or simply an outlier node. However, identifying which classes a node is likely to be indistinguishable is necessary to explicitly adjust the margin of confusing classes. Thus, we suggest Anomalous Distribution-aware Margin (ADM), which complementarily adjusts the target class margin according to the relative closeness to the target class compared to the self class (the class of a given node) in NLD space. Since discriminating two classes is more difficult as NLDs of two classes are closer, we design ADM to be sensitive to the distance between the target class and the self class.

The goal for ADM is to decrease the target class margin more intensively as a given node is closer to the target class compared to the self class in NLD space. Even though there are various methods to compute the relative distance, we calculate the relative distance as in Figure~\ref{fig:concept_figure} (c). Specifically, let us define the angle between the segment between self class-averaged NLD and node NLD, and the segment between self class-averaged NLD and target class-averaged NLD as $A_{v,t}$. Then, following the low of cosine, we can compute $\cos A_{v,t}$ for given node $v$ as:
\begin{equation}~\label{eq:cos}
    \resizebox{1.\linewidth}{!}{
    $\cos A_{v,t} = \frac{{\text{JS}\left(\mathcal{D}_{v,:},\mathcal{C}_{y_v,:}\right)}^2+{\text{JS}\left(\mathcal{C}_{t,:},\mathcal{C}_{y_v,:}\right)}^2-{\text{JS}\left(\mathcal{D}_{v,:},\mathcal{C}_{t,:}\right)}^2}{2{\text{JS}\left(\mathcal{D}_{v,:},\mathcal{C}_{y_v,:}\right)}{\text{JS}\left(\mathcal{C}_{t,:},\mathcal{C}_{y_v,:}\right)}},
    $}    
\end{equation}
where JS is Jensen-Shannon Divergence. Then, ADM is computed as:
\begin{align}
    m^{ADM}_{v,t}= -\frac{\text{JS}\left(\mathcal{D}_{v,:},\mathcal{C}_{y_v,:}\right)\cos A_{v,t}}{\text{JS}\left(\mathcal{C}_{t,:},\mathcal{C}_{y_v,:}\right)}. \label{eq:adm}
\end{align}
Note that ADM can also increase the margin if a node is clearly distinguishable with the given local topology.

\subsection{Class-wise Temperature for Unlabeled Nodes}~\label{subsec:cls-temperature}
Until this point, we have assumed that the label information for neighbors of labeled nodes can be accessible during calculating the NLD $\mathcal{D}$ and Class-wise Connectivity Matrix $\mathcal{C}$. However, in most node classification scenarios, label information is unknown except for a small set of labeled node set $V^{L}$. Therefore, to estimate the class information required when obtaining $\mathcal{D}$ and $\mathcal{C}$, we exploit the model prediction of the model being trained.

To refine the model predictions, we introduce the class-wise temperature strategy. A similar concept is also adopted in the computer vision domain~\citep{ride}, but we use the reverse direction of temperature compared to an existing method. As investigated in previous research~\citep{graphens}, GNNs are prone to overfit to minor class instances in class-imbalanced settings. This issue can bias the neighbors of minor nodes to be over-confident as minor classes. Inspired by this problem, we assign the temperature $T_k$ to logits of each class $k$ based on its quantity $N_k$. When $N_k$ is small, the logits of class $k$ are scaled by a large $T_k$, so the model predictions become more accurate for minor classes. We only use class-wise temperature $T_k$ to obtain the label of neighbors and not for training. The temperature $T_k$ of class $k$ is derived as:
\begin{gather} 
    \pi_{k}=\delta \cdot \frac{N_{k}}{\frac{1}{\vert\mathcal{Y}\vert} \sum_{s \in \vert\mathcal{Y}\vert} N_{s}}+(1-\delta) \nonumber \\
    T_{k}=\frac{1}{\phi\left(\pi_{k}+1-\max _{j} \pi_{j}\right)}, \label{eq:classtemp}
\end{gather}
where $\phi$ is a hyperparameter and $\delta$ is a parameter that determines the sensitivity to imbalance ratio. We fix $\delta$ to 0.4 for all experiments. 

\section{Experiments}
\begin{table*}[t]
\caption{\small Experimental results of our algorithm TAM and other baselines on three class-imbalanced node classification benchmark datasets (homophilous graphs). We report averaged balanced accuracy (bAcc.) and F1-score with the standard errors for 10 repetitions on three representative GNN architectures.}
\begin{center}
\begin{scriptsize}
\setlength{\columnsep}{1pt}%
\begin{adjustbox}{width=0.9\linewidth}
\begin{tabular}{@{\extracolsep{1pt}}rlcc|cc|cc@{}}
\toprule
 & \multirow{1}{*}{\textbf{Dataset}} & \multicolumn{2}{c}{Cora} & \multicolumn{2}{c}{CiteSeer} & \multicolumn{2}{c}{PubMed}  \\ 
\cline{2-8}\rule{0pt}{2.2ex}
& \textbf{Imbalance Ratio ($\rho=10$)} & bAcc. & F1 & bAcc. & F1 & bAcc. & F1 \\
\cline{2-8}
\rule{0pt}{2.5ex}  
\multirow{11}{*}{\rotatebox{90}{GCN}} & Cross Entropy 
                    & 60.95 \tiny{$\pm 1.22$} & 59.30 \tiny{$\pm 1.66$}
                    & 38.21 \tiny{$\pm 1.12$}& 29.40 \tiny{$\pm 1.97$}
                    & 65.21 \tiny{$\pm 1.40$}& 55.43 \tiny{$\pm 2.79$}
                    \\
                    \rule{0pt}{2ex}
                     & Re-Weight
                     & 65.52 \tiny{$\pm 0.84$}& 65.54 \tiny{$\pm 1.20$}
                     & 44.52 \tiny{$\pm 1.22$}& 38.85 \tiny{$\pm 1.62$}
                     & 70.17 \tiny{$\pm 1.25$}& 66.37 \tiny{$\pm 1.73$}
                     \\
                    
                     & PC Softmax 
                     & 67.79 \tiny{$\pm 0.92$}& 67.39 \tiny{$\pm 1.08$}
                     & 49.81 \tiny{$\pm 1.12$}& 45.55 \tiny{$\pm 1.26$}
                     & 70.20 \tiny{$\pm 0.60$}& 68.83 \tiny{$\pm 0.73$}
                     \\
                     
                     & DR-GCN 
                     & 60.17 \tiny{$\pm 0.83$}& 59.31 \tiny{$\pm 0.97$}
                     & 42.64 \tiny{$\pm 0.75$}& 38.22 \tiny{$\pm 1.22$}
                     & 65.51 \tiny{$\pm 0.81$}& 64.95 \tiny{$\pm 0.53$}
                     \\

                     & GraphSMOTE 
                     & 66.29 \tiny{$\pm 0.93$}& 66.30 \tiny{$\pm 1.25$}
                     & 44.40 \tiny{$\pm 1.27$}& 39.10 \tiny{$\pm 1.78$}
                     & 68.51 \tiny{$\pm 1.14$}& 62.63 \tiny{$\pm 2.39$}
                     \\
                     \cline{2-8}
                     
                     & BalancedSoftmax 
                     & 68.46 \tiny{$\pm 0.67$}& 68.41 \tiny{$\pm 0.80$}
                     & 53.70 \tiny{$\pm 1.40$}& 50.73 \tiny{$\pm 1.64$}
                     & 72.97 \tiny{$\pm 0.80$}& 70.80 \tiny{$\pm 1.11$}
                     \\
                     & + \textbf{TAM}
                     & 69.90 \tiny{$\pm 0.73$}& 69.89 \tiny{$\pm 0.89$}
                     & 55.54 \tiny{$\pm 1.40$}& 54.18 \tiny{$\pm 1.69$}
                     & \textbf{74.13} \tiny{$\pm 0.70$}& \textbf{73.27} \tiny{$\pm 0.67$}
                     \\
                     \cdashline{2-8}
                     & ReNode 
                     & 67.61 \tiny{$\pm 0.77$}& 67.27 \tiny{$\pm 0.91$}
                     & 47.78 \tiny{$\pm 1.67$}& 42.51 \tiny{$\pm 2.30$}
                     & 71.59 \tiny{$\pm 1.70$}& 66.56 \tiny{$\pm 2.90$}
                     \\
                     & + \textbf{TAM} 
                     & 67.18 \tiny{$\pm 1.32$}& 67.39 \tiny{$\pm 1.62$}
                     & 48.36 \tiny{$\pm 1.63$}& 42.48 \tiny{$\pm 2.10$}
                     & 71.00 \tiny{$\pm 1.86$}& 67.18 \tiny{$\pm 3.42$}
                     \\
                     \cdashline{2-8}
                     & GraphENS 
                     & 70.31 \tiny{$\pm 0.51$}& 70.30 \tiny{$\pm 0.65$} 
                     & 55.42 \tiny{$\pm 1.74$}& 53.85 \tiny{$\pm 2.00$}
                     & 71.89 \tiny{$\pm 0.80$}& 71.07 \tiny{$\pm 0.66$}
                     \\
                     & + \textbf{TAM}
                     & \textbf{71.52} \tiny{$\pm 0.30$}& \textbf{71.71} \tiny{$\pm 0.45$}
                     & \textbf{57.47} \tiny{$\pm 1.56$}& \textbf{56.23} \tiny{$\pm 1.87$}
                     & 74.01 \tiny{$\pm 0.73$}& 72.41 \tiny{$\pm 0.94$}
                     \\
\cline{2-8}
\noalign{\vskip\doublerulesep
         \vskip-\arrayrulewidth} \cline{2-8}
\rule{0pt}{2.5ex}  
\multirow{11}{*}{\rotatebox{90}{GAT}} & Cross Entropy 
                    & 60.82 \tiny{$\pm 1.27$}& 59.56 \tiny{$\pm 1.75$}
                   
                    & 41.16 \tiny{$\pm 1.49$}& 33.71 \tiny{$\pm 2.02$}
                    
                    & 63.97 \tiny{$\pm 1.21$}& 54.59 \tiny{$\pm 2.42$}
                    \\
                    \rule{0pt}{2ex}
                     & Re-Weight 
                     & 66.72 \tiny{$\pm 0.80$}& 66.52 \tiny{$\pm 1.06$}
                     
                     & 45.59 \tiny{$\pm 1.73$}& 39.43 \tiny{$\pm 2.03$}
                    
                     & 69.13 \tiny{$\pm 1.25$}& 64.81 \tiny{$\pm 1.70$}
                     \\
                    
                     & PC Softmax 
                     & 67.02 \tiny{$\pm 0.65$}& 66.57 \tiny{$\pm 0.89$}
                     
                     & 50.70 \tiny{$\pm 1.73$}& 47.14 \tiny{$\pm 1.85$}
                     
                     & 72.20 \tiny{$\pm 0.49$}& 70.95 \tiny{$\pm 0.82$}
                     \\
                     
                     & DR-GCN 
                     & 59.30 \tiny{$\pm 0.76$}& 57.79 \tiny{$\pm 1.03$}
                     
                     & 44.04 \tiny{$\pm 1.26$}& 39.44 \tiny{$\pm 1.76$}
                    
                     & 69.56 \tiny{$\pm 1.01$}& 68.49 \tiny{$\pm 0.71$}
                     \\

                     & GraphSMOTE 
                     & 66.08 \tiny{$\pm 1.37$}& 64.92 \tiny{$\pm 1.66$}
                    
                     & 45.79 \tiny{$\pm 1.42$}& 39.92 \tiny{$\pm 2.11$}
                    
                     & 67.86 \tiny{$\pm 1.58$}& 61.96 \tiny{$\pm 2.61$}
                     \\
                     
                     \cline{2-8}
                     
                     & BalancedSoftmax 
                     & 67.79 \tiny{$\pm 0.54$}& 67.73 \tiny{$\pm 0.68$}
                     
                     & 52.83 \tiny{$\pm 1.25$}& 49.96 \tiny{$\pm 1.69$}
                     
                     & 72.56 \tiny{$\pm 0.66$}& 69.90 \tiny{$\pm 1.13$}
                     \\
                     & + \textbf{TAM}
                     & 69.00 \tiny{$\pm 0.62$}& 69.25 \tiny{$\pm 0.64$}
                    
                     & \textbf{56.32} \tiny{$\pm 1.65$}& \textbf{54.99} \tiny{$\pm 2.10$}
                     
                     & 73.37 \tiny{$\pm 0.76$}& \textbf{72.60} \tiny{$\pm 0.89$}
                     \\
                     \cdashline{2-8}
                     & ReNode 
                     & 68.34 \tiny{$\pm 1.25$}& 68.59 \tiny{$\pm 1.51$}
                     
                     & 48.99 \tiny{$\pm 1.69$}& 43.90 \tiny{$\pm 2.15$}
                    
                     & 67.55 \tiny{$\pm 2.20$}& 64.46 \tiny{$\pm 2.56$}
                     \\
                     & + \textbf{TAM}
                     & 68.39 \tiny{$\pm 1.15$}& 68.69 \tiny{$\pm 1.36$}
                     
                     & 48.81 \tiny{$\pm 1.26$}& 44.40 \tiny{$\pm 1.91$}
                    
                     & 69.00 \tiny{$\pm 2.39$}& 67.46 \tiny{$\pm 2.66$}
                     \\
                     \cdashline{2-8}
                     & GraphENS 
                     & \textbf{70.45} \tiny{$\pm 0.49$}& 69.84 \tiny{$\pm 0.53$}
                     
                     & 52.35 \tiny{$\pm 1.57$}& 49.35 \tiny{$\pm 2.31$}
                    
                     & 71.99 \tiny{$\pm 0.72$}& 70.59 \tiny{$\pm 0.85$}
                     \\
                     & + \textbf{TAM}
                     & 70.00 \tiny{$\pm 0.60$}& \textbf{69.93} \tiny{$\pm 0.76$}
                     
                     & 55.86 \tiny{$\pm 1.48$}& 53.85 \tiny{$\pm 1.98$}
                     
                     & \textbf{73.42} \tiny{$\pm 0.77$}& 71.95 \tiny{$\pm 1.01$}
                     \\
\cline{2-8}
\noalign{\vskip\doublerulesep
         \vskip-\arrayrulewidth} \cline{2-8}
\rule{0pt}{2.5ex}  
\multirow{11}{*}{\rotatebox{90}{SAGE}} & Cross Entropy 
                    & 60.41 \tiny{$\pm 1.09$}& 58.57 \tiny{$\pm 1.34$}
                    & 44.41 \tiny{$\pm 1.21$}& 38.20 \tiny{$\pm 1.68$}
                    & 67.34 \tiny{$\pm 0.93$}& 62.92 \tiny{$\pm 1.38$}
                    \\
                    \rule{0pt}{2ex}
                     & Re-Weight
                     & 63.76 \tiny{$\pm 0.98$}& 63.46 \tiny{$\pm 1.22$}
                     & 46.64 \tiny{$\pm 1.92$}& 41.38 \tiny{$\pm 2.76$}
                     & 69.03 \tiny{$\pm 1.17$}& 64.01 \tiny{$\pm 2.18$}
                     \\
                    
                     & PC Softmax 
                     & 64.03 \tiny{$\pm 0.81$}& 63.73 \tiny{$\pm 0.99$}
                     & 50.14 \tiny{$\pm 1.89$}& 47.38 \tiny{$\pm 2.13$}
                     & 71.39 \tiny{$\pm 0.84$}& 70.25 \tiny{$\pm 1.02$}
                     \\
                     
                     & DR-GCN 
                     & 61.05 \tiny{$\pm 1.17$}& 60.17 \tiny{$\pm 1.23$}
                     & 46.00 \tiny{$\pm 0.93$}& 47.73 \tiny{$\pm 1.12$}
                     & 69.23 \tiny{$\pm 0.68$}& 67.35 \tiny{$\pm 0.90$}
                     \\

                     & GraphSMOTE
                     & 61.75 \tiny{$\pm 0.07$}& 60.90 \tiny{$\pm 1.22$}
                     & 42.51 \tiny{$\pm 1.54$}& 34.93 \tiny{$\pm 1.67$}
                     & 66.11 \tiny{$\pm 1.12$}& 61.17 \tiny{$\pm 2.10$}
                     \\
                     
                     \cline{2-8}
                     
                     & BalancedSoftmax
                     & 66.10 \tiny{$\pm 0.54$}& 66.26 \tiny{$\pm 0.63$}
                     & 54.18 \tiny{$\pm 1.79$}& 52.67 \tiny{$\pm 1.96$}
                     & 70.32 \tiny{$\pm 0.92$}& 68.81 \tiny{$\pm 0.99$}
                     \\
                     & + \textbf{TAM}
                     & 68.01 \tiny{$\pm 0.56$}& 68.14 \tiny{$\pm 0.58$}
                     & \textbf{55.47} \tiny{$\pm 1.33$}& \textbf{54.87} \tiny{$\pm 1.33$}
                     & \textbf{72.91} \tiny{$\pm 0.62$}& \textbf{72.61} \tiny{$\pm 0.68$}
                     \\
                     \cdashline{2-8}
                     & ReNode 
                     & 65.18 \tiny{$\pm 0.96$}& 65.46 \tiny{$\pm 1.28$}
                     & 48.58 \tiny{$\pm 1.87$}& 43.62 \tiny{$\pm 2.17$}
                     & 69.58 \tiny{$\pm 1.12$}& 67.21 \tiny{$\pm 1.72$}
                     \\
                     & + \textbf{TAM}
                     & 65.54 \tiny{$\pm 1.02$}& 65.96 \tiny{$\pm 1.30$}
                     & 49.53 \tiny{$\pm 1.94$}& 45.96 \tiny{$\pm 2.52$}
                     & 69.96 \tiny{$\pm 1.40$}& 66.34 \tiny{$\pm 2.53$}
                     \\
                     \cdashline{2-8}
                     & GraphENS 
                     & 68.65 \tiny{$\pm 0.49$}& 68.79 \tiny{$\pm 0.21$}
                     & 53.43 \tiny{$\pm 1.29$}& 51.70 \tiny{$\pm 1.46$}
                     & 70.45 \tiny{$\pm 0.82$}& 68.96 \tiny{$\pm 1.34$}
                     \\
                     & + \textbf{TAM}
                     & \textbf{69.12} \tiny{$\pm 0.81$}& \textbf{69.16} \tiny{$\pm 0.87$}
                     & 55.43 \tiny{$\pm 1.32$}& 53.82 \tiny{$\pm 1.49$}
                     & 72.31 \tiny{$\pm 1.05$}& 71.21 \tiny{$\pm 1.31$}
                     \\
\bottomrule
\end{tabular}
 \end{adjustbox}
\end{scriptsize}
\end{center}
\label{tb:main_homo}
\vspace{-0.05in}
\end{table*}
\subsection{Experimental Settings}
\paragraph{Datasets} To show the effectiveness of our algorithm on both homophilous and heterophilous graphs, we evaluate our method on homophilous graphs: Cora, CiteSeer, and PubMed~\citep{DBLP:journals/aim/SenNBGGE08}, and heterophilous graphs: Wisconsin\footnote{http://www.cs.cmu.edu/afs/cs.cmu.edu/project/theo-11/www/wwkb/}, Chameleon, and Squirrel~\citep{wikipedia}. We utilize the splits used in \citet{public_split} for Cora, CiteSeer, and PubMed, and in \citet{geomgcn} for Wisconsin, Chameleon, and Squirrel. To construct class-imbalanced datasets, we adopt the step imbalance method following \citet{graphsmote,graphens}. Specifically, we select minor classes as half the number of classes ($\vert\mathcal{Y}\vert /2$) and alter labeled nodes of minor classes to unlabeled nodes randomly until the number of nodes in each minor class equals the ratio of the number of major nodes in the most frequent class to imbalance ratio ($\frac{\max_{k\in\mathcal{Y}}N_k}{\rho}$). In this paper, we adopt imbalance ratios of five and ten. For Wisconsin, we do not modify the number of nodes in each class since the train splits in \citet{geomgcn} is already highly imbalanced (11.63). The detailed experimental settings such as evaluation protocol and implementation details of our algorithm are described in Appendix~\ref{appen:experiments}.
\begin{table*}[t]
\caption{\small Experimental results of our algorithm TAM and other baselines on three class-imbalanced node classification benchmark datasets (heterophilous graphs). We report averaged balanced accuracy (bAcc.) and F1-score with the standard errors for 10 repetitions on three representative GNN architectures.}
\begin{center}
\begin{scriptsize}
\setlength{\columnsep}{1pt}%
\begin{adjustbox}{width=0.88\linewidth}
\begin{tabular}{@{\extracolsep{1pt}}rlcc|cc|cc@{}}
\toprule
 & \multirow{1}{*}{\textbf{Dataset}} & \multicolumn{2}{c}{Chameleon} & \multicolumn{2}{c}{Squirrel} & \multicolumn{2}{c}{Wisconsin}  \\ 
\cline{2-8}\rule{0pt}{2.2ex}
& \multirow{2}{*}{\textbf{Imbalance Ratio}} &  \multicolumn{2}{c|}{($\rho=5$)}  & \multicolumn{2}{c|}{($\rho=5$)} &  \multicolumn{2}{c}{($\rho=11.63$)}  \\ 
& & bAcc. & F1  & bAcc. & F1 & bAcc. & F1 \\
\cline{2-8}
\rule{0pt}{2.5ex}  
\multirow{11}{*}{\rotatebox{90}{GCN}} & Cross Entropy & 33.21 \tiny{$\pm 0.88$}& 31.74 \tiny{$\pm 0.85$}
                    
                    & 24.06 \tiny{$\pm 0.43$}& 20.32 \tiny{$\pm 0.59$}
                    
                    & 29.73 \tiny{$\pm 1.29$}& 27.51 \tiny{$\pm 1.45$}
                    \\
                    \rule{0pt}{2ex}
                     & Re-Weight & 37.85 \tiny{$\pm 0.95$}& 37.46 \tiny{$\pm 0.95$}
                    
                     & 27.40 \tiny{$\pm 0.52$}& 26.76 \tiny{$\pm 0.42$}
                     
                     & 44.13 \tiny{$\pm 3.08$}& 40.74 \tiny{$\pm 3.27$}
                     \\
                    
                     & PC Softmax & 37.98 \tiny{$\pm 0.83$}& 36.55 \tiny{$\pm 0.88$}
                     
                     & 27.37 \tiny{$\pm 0.33$}& 26.67 \tiny{$\pm 0.27$}
                     
                     & 30.90 \tiny{$\pm 3.10$}& 28.15 \tiny{$\pm 2.16$}
                     \\
                     
                     & DR-GCN & 34.12 \tiny{$\pm 0.89$}& 31.78 \tiny{$\pm 1.02$}
                     
                     & 24.67 \tiny{$\pm 0.39$}& 19.54 \tiny{$\pm 0.71$}
                     
                     & 29.44 \tiny{$\pm 1.36$}& 27.08 \tiny{$\pm 1.37$}
                     \\

                     & GraphENS & 41.13 \tiny{$\pm 0.59$}& 39.61 \tiny{$\pm 0.77$}
                    
                     & 26.79 \tiny{$\pm 0.43$}& 26.35 \tiny{$\pm 0.41$}
                     & 44.09 \tiny{$\pm 2.96$}& 40.86 \tiny{$\pm 3.29$}
                     \\
                     
                     \cline{2-8}
                     
                     & BalancedSoftmax & 38.33 \tiny{$\pm 0.73$}& 37.54 \tiny{$\pm 0.68$}
                    
                     & 27.86 \tiny{$\pm 0.42$}& 27.04 \tiny{$\pm 0.35$}
                    
                     & 31.51 \tiny{$\pm 2.28$}& 28.82 \tiny{$\pm 1.99$}
                     \\
                     & + \textbf{TAM} & 41.48 \tiny{$\pm 0.93$}& 40.43 \tiny{$\pm 1.02$}
                    
                     & 28.67 \tiny{$\pm 0.54$}& 27.84 \tiny{$\pm 0.45$}
                     
                     & 35.97 \tiny{$\pm 3.68$}& 31.41 \tiny{$\pm 2.16$}
                     \\
                     \cdashline{2-8}
                     & ReNode & 37.43 \tiny{$\pm 0.90$}& 36.75 \tiny{$\pm 0.89$}
                    
                     & 28.38 \tiny{$\pm 0.46$}& 27.81 \tiny{$\pm 0.44$}
                   
                     & 36.85 \tiny{$\pm 2.14$}& 33.30 \tiny{$\pm 1.79$}
                     \\
                     & + \textbf{TAM} & 40.28 \tiny{$\pm 0.85$}& 39.27 \tiny{$\pm 0.80$}
                     
                     & 28.19 \tiny{$\pm 0.36$}& 27.55 \tiny{$\pm 0.39$}
                    
                     & 36.28 \tiny{$\pm 2.87$}& 34.10 \tiny{$\pm 2.38$}
                     \\
                     \cdashline{2-8}
                     & GraphSMOTE & 42.65 \tiny{$\pm 0.59$}& 41.56 \tiny{$\pm 0.53$}
                     
                     & 28.29 \tiny{$\pm 0.60$}& 27.89 \tiny{$\pm 0.61$}
                    
                     & 45.36 \tiny{$\pm 4.21$}& \textbf{40.91} \tiny{$\pm 4.39$}
                     \\
                     & + \textbf{TAM} & \textbf{42.77} \tiny{$\pm 0.62$}& \textbf{41.78} \tiny{$\pm 0.62$}
                   
                     & \textbf{29.18} \tiny{$\pm 0.46$}& \textbf{28.84} \tiny{$\pm 0.44$}
                 
                     & \textbf{45.59} \tiny{$\pm 3.73$}& 40.76 \tiny{$\pm 4.01$}
                     \\
\cline{2-8}
\noalign{\vskip\doublerulesep
         \vskip-\arrayrulewidth} \cline{2-8}
\rule{0pt}{2.5ex}  
\multirow{11}{*}{\rotatebox{90}{GAT}} & Cross Entropy & 34.33 \tiny{$\pm 0.74$}& 31.54 \tiny{$\pm 0.95$}
                    
                    & 24.89 \tiny{$\pm 0.37$}& 21.33 \tiny{$\pm 0.52$}
                  
                    & 32.15 \tiny{$\pm 2.72$}& 30.92 \tiny{$\pm 2.76$}
                    \\
                    \rule{0pt}{2ex}
                     & Re-Weight & 39.63 \tiny{$\pm 0.49$}& 39.08 \tiny{$\pm 0.50$}
                    
                     & 26.49 \tiny{$\pm 0.41$}& 25.92 \tiny{$\pm 0.41$}
                   
                     & 42.15 \tiny{$\pm 2.33$}& 37.66 \tiny{$\pm 2.27$}
                     \\
                    
                     & PC Softmax & 41.47 \tiny{$\pm 0.78$}& 40.51 \tiny{$\pm 0.89$}
                     
                     & 27.31 \tiny{$\pm 0.51$}& 26.74 \tiny{$\pm 0.50$}
                    
                     & 41.89 \tiny{$\pm 3.95$}& 38.03 \tiny{$\pm 3.35$}
                     \\
                     
                     & DR-GCN & 36.85 \tiny{$\pm 0.77$}& 34.61 \tiny{$\pm 0.62$}
                    
                     & 25.40 \tiny{$\pm 0.43$}& 22.83 \tiny{$\pm 0.59$}
                   
                     & 33.93 \tiny{$\pm 2.34$}& 31.75 \tiny{$\pm 2.50$}
                     \\

                     & GraphENS & 40.66 \tiny{$\pm 1.13$}& 39.49 \tiny{$\pm 1.10$}
                    
                     & 26.87 \tiny{$\pm 0.43$}& 26.78 \tiny{$\pm 0.41$}  
                  
                     & 40.93 \tiny{$\pm 2.78$}& 37.43 \tiny{$\pm 2.74$}
                     \\
                     
                     \cline{2-8}
                     
                     & BalancedSoftmax & 41.47 \tiny{$\pm 0.71$}& 40.52 \tiny{$\pm 0.78$}
                     
                     & 26.66 \tiny{$\pm 0.39$}& 25.97 \tiny{$\pm 0.35$}
                    
                     & 41.20 \tiny{$\pm 3.08$}& 37.93 \tiny{$\pm 2.99$}
                     \\
                     & + \textbf{TAM} & 42.56 \tiny{$\pm 0.59$}& 41.40 \tiny{$\pm 0.74$}
                     
                     & 27.75 \tiny{$\pm 0.44$}& 27.23 \tiny{$\pm 0.45$}
                    
                     & \textbf{48.44} \tiny{$\pm 3.32$}& \textbf{43.71} \tiny{$\pm 2.91$}
                     \\
                     \cdashline{2-8}
                     & ReNode & 40.41 \tiny{$\pm 0.56$}& 39.85 \tiny{$\pm 0.60$}
                     
                     & 26.89 \tiny{$\pm 0.45$}& 26.40 \tiny{$\pm 0.46$}
                    
                     & 40.88 \tiny{$\pm 2.84$}& 37.13 \tiny{$\pm 2.74$}
                     \\
                     
                     & + \textbf{TAM} & 41.53 \tiny{$\pm 0.35$}& 40.76 \tiny{$\pm 0.50$}
                     
                     & 26.53 \tiny{$\pm 0.40$}& 26.00 \tiny{$\pm 0.42$}
                     
                     & 46.64 \tiny{$\pm 3.35$}& 41.60 \tiny{$\pm 3.02$}
                    
                     \\
                     \cdashline{2-8}
                     & GraphSMOTE & 42.27 \tiny{$\pm 0.51$}& 41.43 \tiny{$\pm 0.54$}
                   
                     & 28.17 \tiny{$\pm 0.56$}& 27.38 \tiny{$\pm 0.66$}
                     
                     & 40.77 \tiny{$\pm 2.24$}& 38.96 \tiny{$\pm 2.48$}
                     \\
                     & + \textbf{TAM} & \textbf{42.83} \tiny{$\pm 0.82$}& \textbf{42.26} \tiny{$\pm 0.83$}
                    
                     & \textbf{28.44} \tiny{$\pm 0.33$} & \textbf{28.02} \tiny{$\pm 0.37$}
                     
                     & 41.82 \tiny{$\pm 2.94$}& 38.23 \tiny{$\pm 3.13$}
                     \\
\cline{2-8}
\noalign{\vskip\doublerulesep
         \vskip-\arrayrulewidth} \cline{2-8}
\rule{0pt}{2.5ex}  
\multirow{11}{*}{\rotatebox{90}{SAGE}} & Cross Entropy & 35.76 \tiny{$\pm 0.57$}& 33.55 \tiny{$\pm 0.68$}
                    
                    & 27.59 \tiny{$\pm 0.31$}& 25.87 \tiny{$\pm 0.14$}
                    
                    & 68.76 \tiny{$\pm 3.57$}& 64.16 \tiny{$\pm 3.26$}
                    \\
                    \rule{0pt}{2ex}
                     & Re-Weight & 40.85 \tiny{$\pm 0.69$}& 40.40 \tiny{$\pm 0.71$}
                     
                     & 29.88 \tiny{$\pm 0.48$}& 27.59 \tiny{$\pm 0.42$}
                   
                     & 68.13 \tiny{$\pm 3.19$}& 63.45 \tiny{$\pm 2.27$}
                     \\
                    
                     & PC Softmax & 42.90 \tiny{$\pm 0.85$}& 42.34 \tiny{$\pm 0.87$}
                    
                     & 30.54 \tiny{$\pm 0.62$}& 29.41 \tiny{$\pm 0.65$}
                   
                     & 70.57 \tiny{$\pm 3.34$}& 67.13 \tiny{$\pm 2.91$}
                     \\
                     
                     & DR-GCN & 39.58 \tiny{$\pm 0.58$}& 38.37 \tiny{$\pm 0.72$}
                     
                     & 28.78 \tiny{$\pm 0.50$}& 25.01 \tiny{$\pm 0.70$}
                    
                     & 69.30 \tiny{$\pm 1.99$}& 64.60 \tiny{$\pm 2.00$}
                  
                     \\

                     & GraphENS & 37.77 \tiny{$\pm 0.69$}& 37.36 \tiny{$\pm 0.67$}
                     & 25.31 \tiny{$\pm 0.45$}& 25.15 \tiny{$\pm 0.40$}
                     
                     & 66.23 \tiny{$\pm 3.17$}& 60.89 \tiny{$\pm 2.97$}
                   
                     \\
                     
                     \cline{2-8}
                     
                     & BalancedSoftmax & 43.03 \tiny{$\pm 0.98$}& 42.40 \tiny{$\pm 0.96$}
                     
                     & 30.29 \tiny{$\pm 0.48$}& 29.37 \tiny{$\pm 0.44$}
                     & 67.50 \tiny{$\pm 2.47$}& 63.95 \tiny{$\pm 2.43$}
                     \\
                     & + \textbf{TAM} & \textbf{43.77} \tiny{$\pm 0.90$}& \textbf{42.95} \tiny{$\pm 0.90$}
                    
                     & \textbf{30.70} \tiny{$\pm 0.59$}& \textbf{29.82} \tiny{$\pm 0.53$}
                    
                     & 68.62 \tiny{$\pm 3.47$}& 65.23 \tiny{$\pm 2.97$}
                     \\
                     \cdashline{2-8}
                     & ReNode & 40.74 \tiny{$\pm 0.75$}& 40.45 \tiny{$\pm 0.77$}
                     & 29.75 \tiny{$\pm 0.47$}& 28.49 \tiny{$\pm 0.51$}
                     & \textbf{72.52} \tiny{$\pm 2.13$}& \textbf{69.15} \tiny{$\pm 3.18$}
                     \\
                     & + \textbf{TAM} & 41.45 \tiny{$\pm 0.86$}& 41.00 \tiny{$\pm 0.86$}
                    
                     & 29.79 \tiny{$\pm 0.39$} & 28.75 \tiny{$\pm 0.39$}
                    
                     & 70.97 \tiny{$\pm 2.05$}& 66.95 \tiny{$\pm 2.62$}
                     \\
                     \cdashline{2-8}
                     & GraphSMOTE & 34.43 \tiny{$\pm 0.86$}& 31.16 \tiny{$\pm 1.21$}
                     
                     & 26.26 \tiny{$\pm 0.31$}& 23.73 \tiny{$\pm 0.30$}
                     
                     & 65.14 \tiny{$\pm 3.84$}& 62.53 \tiny{$\pm 3.40$}
                     \\
                     & + \textbf{TAM} & 36.94 \tiny{$\pm 0.88$}& 35.00 \tiny{$\pm 0.93$}
                     & 26.70 \tiny{$\pm 0.48$} & 24.71 \tiny{$\pm 0.41$}
                     & 64.07 \tiny{$\pm 3.15$}& 62.65 \tiny{$\pm 2.96$}
                     \\

\bottomrule

\end{tabular}
\end{adjustbox}
\end{scriptsize}
\end{center}
\label{tb:main_hetero}
\vspace{-0.1in}
\end{table*}
\subsection{Baselines} 
To validate our method, we first select vanilla (cross entropy) and re-weight~\citep{reweight} as baselines. We also adopt competitive baselines in the vision domain and node classification. For the vision domain, Balanced Softmax~\citep{balanced_softmax} and PC Softmax~\citep{pc_softmax} are introduced as the representative algorithms of loss modification and post-hoc correction, respectively. In node classification, we compare our method with DR-GCN~\citep{conditionalgan}, GraphSMOTE~\citep{graphsmote}, and GraphENS~\citep{graphens}. For ReNode~\citep{renode}, we search the best algorithm in each setting among re-weight~\citep{reweight}, focal loss~\citep{focal}, and class-balanced weight~\citep{cbsoftmax}. To show that our algorithm could improve the performance of many imbalance handling methods, we combine our methods with competitive approaches in each domain: Balanced Softmax, ReNode, and GraphENS. Since GraphSMOTE shows superior performance only in heterophilous graphs than GraphENS, we combine our methods with GraphSMOTE rather than GraphENS in Wisconsin, Chameleon, and Squirrel. The implementation details of baselines are suggested in Appendix~\ref{appensub:baselines}.

\subsection{Main Results}~\label{subsec:main_results}
\vspace{-0.25in}
\paragraph{Homophilous graphs}
In Table~\ref{tb:main_homo}, we report the averaged balanced accuracy (bAcc.) and F1 score with standard error for the baselines and ours on three homogeneously-connected citation networks. Existing imbalance approaches integrated with TAM achieve the best performance for all 9 settings (3 datasets with 3 architectures). Since our method can detect the nodes of the topological boundaries and adjust their logits based on ACM module (Section~\ref{subsec:acm}), TAM successfully brings enhancement of imbalance-handling performance on homophilous graphs. We also confirm that TAM consistently improves the performance over various types of imbalance handling strategies such as logit adjustment (Balanced Softmax), node weighting (ReNode), and oversampling (GraphENS). Note that we validate our method on another imbalance ratio ($\rho=5$) and get consistent results. These results are deferred to Appendix~\ref{appensub:homo}.   

\vspace{-0.1in}
\paragraph{Heterophilous graphs}
As shown in Table~\ref{tb:main_hetero}, Baselines equipped with TAM also exhibit superior imbalance-handling performances in most cases for three heterogeneously-connected graphs. The rationale for these results is that TAM could model the class-wise connectivity pattern of heterphilous graphs and identify the outlier nodes for each class using class-averaged statistic matrix $\mathcal{C}$. We also verify that TAM indeed reduces the false positives stemming from abnormally-connected minor nodes (see \textbf{Reducing False Positives} in Section~\ref{subsec:main_results}). Note that we also provide the comparison of our method with other baselines under another imbalance ratio ($\rho=10$). These results are deferred to Appendix~\ref{appensub:hetero}.

\vspace{-0.11in}
\paragraph{Reducing False Positives}
In Section~\ref{sec:problem}, we have observed that false positives on minor classes are highly located near the minor nodes which have anomalous connectivity. To verify that TAM can alleviate this false positive issue, we compute the ratio of false positives on neighbor nodes of anomalous minor nodes using GNNs trained with TAM. As indicated in Figure~\ref{fig:fp_topology} (c) and (d), TAM steadily reduces the ratio of false positives without dependence of imbalance handling approaches on both homophilous and heterophilous graphs.

\subsection{Analysis of TAM}
\paragraph{Ablation study}
To validate each component of our method, we conduct an ablation study on three node classification datasets. First, we compare ACM with ACM without using class-connectivity matrix $\mathcal{C}$ to justify exploiting class connectivity statistics. Specifically `ACM w/o $\mathcal{C}$' solely uses $\frac{D_{v,t}}{D_{v,y_v}}$ to determine the ACM margins. In Table~\ref{tb:ablation}, `ACM w/o ' consistently shows inferior performances compared to our ACM module implying that the class connectivity matrix plays an important role in modeling connectivity patterns.

We also verify our three key modules: ACM, ADM, and Class-wise Temperature (Cls-wise $T_k$). As shown in Table~\ref{tb:ablation}, each component of our method can bring performance improvement by \textit{itself}. From these results, We carefully expect that our individual module contributes to alleviating the adversarial byproduct in the imbalance-handling process. Note that we use Balanced Softmax as a base imbalance-handling approach in the ablation study.

\begin{table}[h]
\vspace{-0.1in}
\center
\caption{\small Ablation Study of our method TAM.}
\begin{scriptsize}
\setlength{\tabcolsep}{2pt} 
\begin{tabular}{c|cccc|c}
\toprule
    \textbf{Modules} & \textbf{ACM w/o} $\mathbf{\mathcal{C}}$ & \textbf{ACM} & \textbf{ADM} & \textbf{Cls-wise $\mathbf{T_k}$} & F1  \\
    \hline
    \multirow{6}{*}{\shortstack[1]{CiteSeer \\ + \\ GCN}} & \grayx & \grayx & \grayx & \grayx  & 50.73 \tiny{$\pm 1.40$} \\
            & \cmark & \grayx & \grayx & \grayx & 50.28 \tiny{$\pm 1.16$}\\
            & \grayx & \cmark & \grayx & \grayx & 53.54 \tiny{$\pm 1.99$} \\
            & \grayx & \grayx & \cmark & \grayx & 51.95 \tiny{$\pm 1.71$} \\
            & \grayx & \cmark & \cmark & \grayx & 54.08 \tiny{$\pm 1.81$} \\
            & \grayx & \cmark & \cmark & \cmark & \textbf{55.54} \tiny{$\pm 1.40$} \\
    \cline{0-5}
    \multirow{6}{*}{\shortstack[1]{PubMed \\ + \\ GCN}} & \grayx & \grayx & \grayx & \grayx  & 70.80 \tiny{$\pm 1.11$} \\
            & \cmark & \grayx & \grayx & \grayx & 71.29 \tiny{$\pm 1.16$}\\
            & \grayx & \cmark & \grayx & \grayx & 72.59 \tiny{$\pm 0.93$} \\
            & \grayx & \grayx & \cmark & \grayx & 71.47 \tiny{$\pm 1.03$} \\
            & \grayx & \cmark & \cmark & \grayx & 72.84 \tiny{$\pm 1.16$} \\
            & \grayx & \cmark & \cmark & \cmark & \textbf{73.27} \tiny{$\pm 2.39$} \\
    \cline{0-5}
    \multirow{6}{*}{\shortstack[1]{chameleon \\ + \\ SAGE}} & \grayx & \grayx & \grayx & \grayx & 42.40 \tiny{$\pm 0.96$}  \\
            & \cmark & \grayx & \grayx & \grayx & 42.34 \tiny{$\pm 0.88$} \\
            & \grayx & \cmark & \grayx & \grayx & 42.53 \tiny{$\pm 1.07$} \\
            & \grayx & \grayx & \cmark & \grayx & 42.75 \tiny{$\pm 1.01$} \\
            & \grayx & \cmark & \cmark & \grayx & \textbf{42.98} \tiny{$\pm 1.00$}\\
            & \grayx & \cmark & \cmark & \cmark & \textbf{42.95} \tiny{$\pm 0.90$}\\
\bottomrule
\end{tabular}
\end{scriptsize}
\label{tb:ablation}
\vspace{-0.1in}
\end{table}

\begin{figure}[h]
    \vspace{-0.05in}
  \centering
   \begin{minipage}[b]{0.235\textwidth}
    \includegraphics[width=\textwidth]{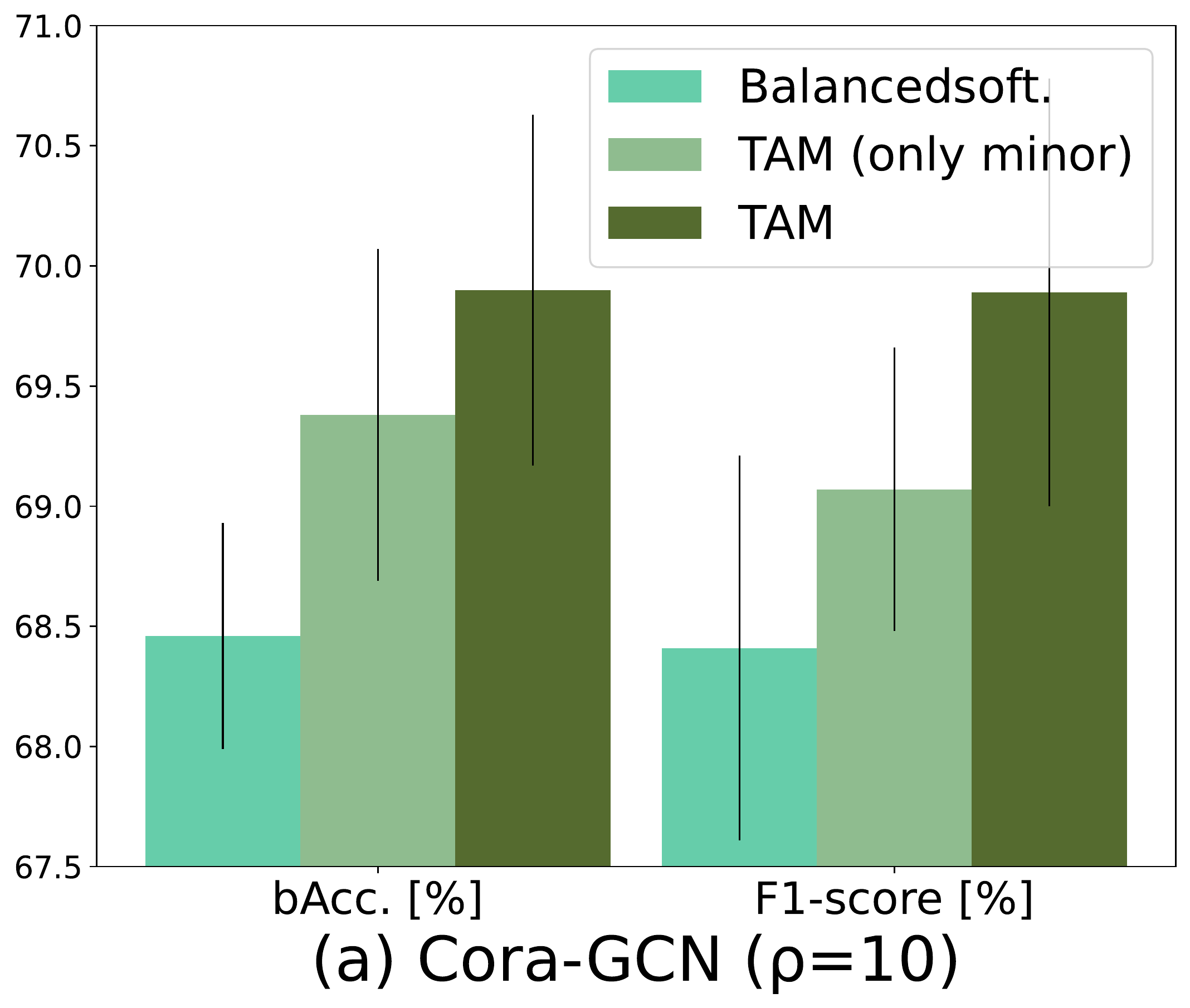}
  \end{minipage}
  \begin{minipage}[b]{0.23\textwidth}
    \includegraphics[width=\textwidth]{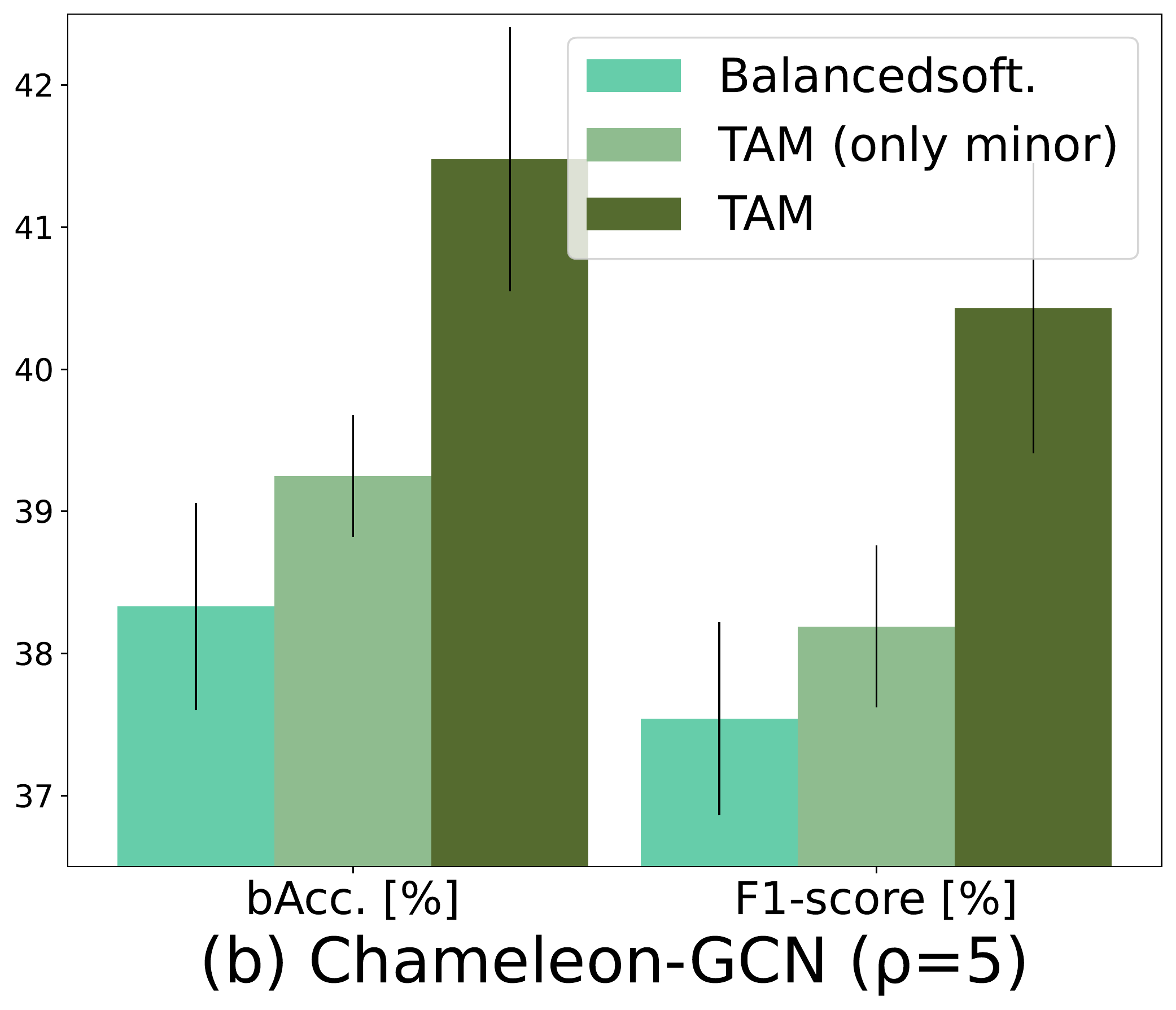}
  \end{minipage}
  \vspace{-0.1in}
  \caption{\small Comparison of balanced accuracy and F1-score. For (a) and (b), performance gains by adjusting all classes compared to adjusting only minor nodes are shown.}
  \vspace{-0.05in}
  \label{fig:only_minor}
\end{figure}

\paragraph{Adjusting both minor and major nodes}
TAM also adjusts the margins for nodes other than minor class nodes (\textit{i}.\textit{e}. major class nodes). The rationale for this design is that, even though the impact of anomalous major nodes is far weaker, they still occur false positives to their neighbors. To validate our principle, we explore the performances of the TAM that \textit{only} regulates the margins of minor nodes on Cora and Chameleon datasets (Figure~\ref{fig:only_minor}). These results support our choice in that adjusting only minor nodes shows sub-optimal performances while it brings considerable improvements compared to baselines.


\begin{figure}[h]
\vspace{-0.05in}
  \centering
  \includegraphics[width=0.48\textwidth]{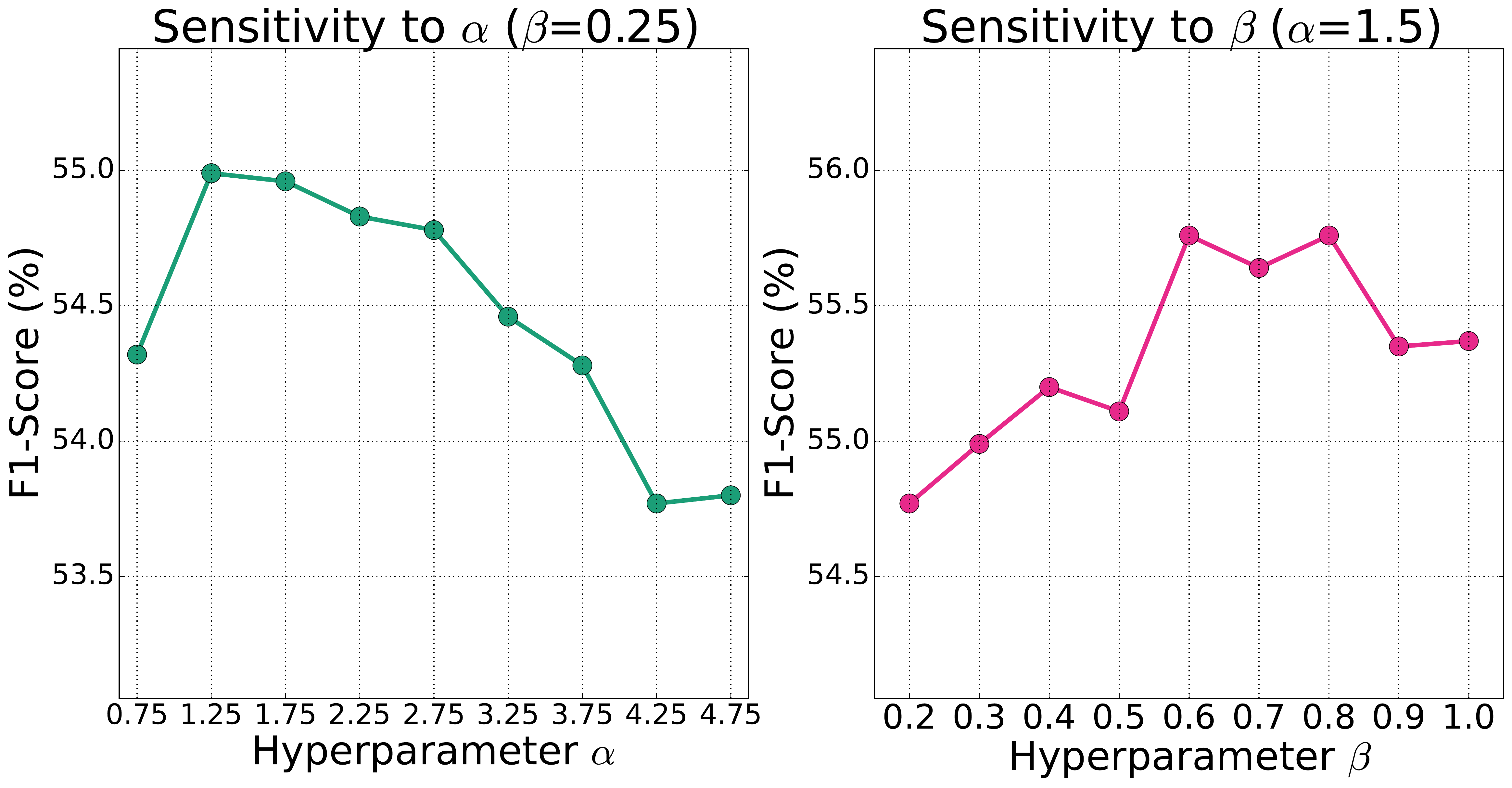}
  \vspace{-0.2in}
  \caption{\small Sensitivity graphs on CiteSeer ($\rho$=10). Green and red graphs show the performance change as ACM intensity $\alpha$ and ADM intensity $\beta$ increases, respectively.}
  \label{fig:sensitivity}
\end{figure}

\paragraph{Sensitivity to hyperparameters $\alpha$ and $\beta$}
The two intensity terms - $\alpha$ and $\beta$ - for ACM and ADM have been introduced in Section~\ref{sec:method}. We investigate the sensitivity of performance to ACM intensity $\alpha$ and ADM intensity $\beta$ in Figure~\ref{fig:sensitivity}. We observe the performance drops when $\alpha$ or $\beta$ have extreme values. We believe that small $\alpha$ or $\beta$ might not sufficiently decrease false positives induced by anomalous nodes. In contrast, large $\alpha$ or $\beta$ would increase false negatives of minor classes.

\section{Related Work}

\paragraph{Imbalance handling in the vision domain} The key objective of solving class-imbalanced problem is to mitigate the bias to major classes induced by the label distribution in the training set. There are four major branches for imbalance handling: re-sampling methods, ensemble approaches, post-hoc correction, and loss modification. Re-sampling methods~\citep{smote,decoupling,bag_tricks,rsg} sample minor class data more frequently or augment minor class diversely. Ensemble approaches~\citep{bbn,lmfe,ride,ace} train multiple head classifiers and collaboratively inference test data with these classifiers. Post-hoc correction algorithms~\citep{decoupling,post_ic,logit_adjustment,pc_softmax} reward minor classes only in the inference. Loss modification methods~\citep{cbsoftmax,logit_adjustment,causal_norm} compensate minor classes in the training phase by assigning more weights on the loss of minor class data~\citep{reweight,focal,cbsoftmax,lcbal} or expanding the margin of minor classes to major classes~\citep{ldam,equalization,balanced_softmax,logit_adjustment,acl,seesaw}. However, it is challenging to directly apply these methods to node classification due to connections between nodes in graphs.

\paragraph{Imbalance handling in node classification} Recently, to utilize topology information in class-imbalanced node classification, several methods~\citep{conditionalgan,rect,graphsmote,pc_gnn,imgagn,renode} are proposed. DR-GCN~\citep{conditionalgan} introduces conditional GAN to generate virtual nodes which are similar to adjacent node features of source nodes. GraphSMOTE~\citep{graphsmote} synthesizes the features of minor nodes by interpolating two minor nodes as SMOTE~\citep{smote} does and determines edges of synthesized nodes with edge predictor. ImGAGN~\citep{imgagn} produces virtual minor nodes by mixing all minor nodes and these virtual nodes connect only to minor nodes according to the generated weight matrix. Since ImGAGN is designed to mainly target binary classification, the extension of ImGAGN to multi-class classification is non-trivial. GraphENS~\citep{graphens} generates diverse minor nodes by mixing minor nodes with (other class) nodes in neighbor distribution level utilizing model prediction and node feature level using saliency map. However, these approaches do not consider the topologies of nodes when rewarding minor classes. ReNode~\citep{renode} reduces loss weights of nearby nodes at topological class boundaries, but it is only effective when graphs are homophilous and does not consider class-pair connectivity. Compared to other imbalance handling methods, TAM reflects class-pair connectivity on logit adjustments and works on both homophilous and heterophilous graphs.
\vspace{-0.05in}

\paragraph{Algorithms for heterophilous graphs} Although our method mainly targets the class-imbalanced problem in node classification, handling heterophilous graphs is related in that we consider connectivity pattern of each class. Since many GNNs are designed under homophily assumption, many GNNs fail in heterophilous graphs. Recently, to overcome this limitation, various algorithms~\citep{mixhop,geomgcn,h2gnn,cpgnn,ugcn,classhomo,dmp} are suggested. CPGNN~\citep{cpgnn} models Compatibility Matrix and conducts propagation with this matrix. DMP~\citep{dmp} determines edge attributes for each edge and propagates node features with these attributes.

\section{Conclusion}

In class-imbalanced node classification, we found that compensating minor nodes, which deviate from class-wise connectivity patterns, are prone to induce false positives cases for major nodes. From this observation, we proposed TAM to adjust margins node-wisely according to the extent of deviation from connectivity patterns. We show that our algorithm consistently improves competitive imbalance handling methods by simply combining TAM on both homophilous and heterophilous graphs with various GNN architectures. 

\section*{Acknowledgements}
This work was supported by Institute of Information \& Communications Technology Planning
\& Evaluation (IITP) grant funded by the Korea government (MSIT) (No.2019-0-00075, Artificial Intelligence Graduate School Program
(KAIST), No.2019-0-01371, Development of brain-inspired AI with human-like intelligence No.2019-0-00075, Artificial Intelligence Innovation Hub, No.2022-0-00713, Meta-learning applicable to real-world problems) and the National Research Foundation
of Korea (NRF) grants (No.2018R1A5A1059921) funded by the Korea government (MSIT). This
work was also supported by Samsung Electronics Co., Ltd (No.IO201214-08133-01).



\bibliography{icml2022}

\begin{thebibliography}{46}
\providecommand{\natexlab}[1]{#1}
\providecommand{\url}[1]{\texttt{#1}}
\expandafter\ifx\csname urlstyle\endcsname\relax
  \providecommand{\doi}[1]{doi: #1}\else
  \providecommand{\doi}{doi: \begingroup \urlstyle{rm}\Url}\fi

\bibitem[Abu-El-Haija et~al.(2019)Abu-El-Haija, Perozzi, Kapoor, Alipourfard,
  Lerman, Harutyunyan, Ver~Steeg, and Galstyan]{mixhop}
Abu-El-Haija, S., Perozzi, B., Kapoor, A., Alipourfard, N., Lerman, K.,
  Harutyunyan, H., Ver~Steeg, G., and Galstyan, A.
\newblock Mixhop: Higher-order graph convolutional architectures via sparsified
  neighborhood mixing.
\newblock In \emph{international conference on machine learning}, pp.\  21--29.
  PMLR, 2019.

\bibitem[Cai et~al.(2021)Cai, Wang, and Hwang]{ace}
Cai, J., Wang, Y., and Hwang, J.-N.
\newblock Ace: Ally complementary experts for solving long-tailed recognition
  in one-shot.
\newblock In \emph{Proceedings of the IEEE/CVF International Conference on
  Computer Vision}, pp.\  112--121, 2021.

\bibitem[Cao et~al.(2019)Cao, Wei, Gaidon, Arechiga, and Ma]{ldam}
Cao, K., Wei, C., Gaidon, A., Arechiga, N., and Ma, T.
\newblock Learning imbalanced datasets with label-distribution-aware margin
  loss.
\newblock \emph{Advances in neural information processing systems}, 32, 2019.

\bibitem[Chawla et~al.(2002)Chawla, Bowyer, Hall, and Kegelmeyer]{smote}
Chawla, N.~V., Bowyer, K.~W., Hall, L.~O., and Kegelmeyer, W.~P.
\newblock {SMOTE:} synthetic minority over-sampling technique.
\newblock \emph{J. Artif. Intell. Res.}, 16:\penalty0 321--357, 2002.

\bibitem[Chen et~al.(2021)Chen, Lin, Zhao, Ren, Li, Zhou, and Sun]{renode}
Chen, D., Lin, Y., Zhao, G., Ren, X., Li, P., Zhou, J., and Sun, X.
\newblock Topology-imbalance learning for semi-supervised node classification.
\newblock \emph{Advances in Neural Information Processing Systems},
  34:\penalty0 29885--29897, 2021.

\bibitem[Chen et~al.(2018)Chen, Ma, and Xiao]{fastgcn}
Chen, J., Ma, T., and Xiao, C.
\newblock Fastgcn: Fast learning with graph convolutional networks via
  importance sampling.
\newblock In \emph{International Conference on Learning Representations}, 2018.

\bibitem[Cui et~al.(2019)Cui, Jia, Lin, Song, and Belongie]{cbsoftmax}
Cui, Y., Jia, M., Lin, T.-Y., Song, Y., and Belongie, S.
\newblock Class-balanced loss based on effective number of samples.
\newblock In \emph{Proceedings of the IEEE/CVF conference on computer vision
  and pattern recognition}, pp.\  9268--9277, 2019.

\bibitem[Hamilton et~al.(2017)Hamilton, Ying, and Leskovec]{sage}
Hamilton, W., Ying, Z., and Leskovec, J.
\newblock Inductive representation learning on large graphs.
\newblock \emph{Advances in neural information processing systems}, 30, 2017.

\bibitem[Hong et~al.(2021)Hong, Han, Choi, Seo, Kim, and Chang]{pc_softmax}
Hong, Y., Han, S., Choi, K., Seo, S., Kim, B., and Chang, B.
\newblock Disentangling label distribution for long-tailed visual recognition.
\newblock In \emph{{IEEE} Conference on Computer Vision and Pattern
  Recognition, {CVPR} 2021, virtual, June 19-25, 2021}, pp.\  6626--6636.
  Computer Vision Foundation / {IEEE}, 2021.

\bibitem[Japkowicz \& Stephen(2002)Japkowicz and Stephen]{reweight}
Japkowicz, N. and Stephen, S.
\newblock The class imbalance problem: {A} systematic study.
\newblock \emph{Intell. Data Anal.}, 6\penalty0 (5):\penalty0 429--449, 2002.

\bibitem[Jin et~al.(2021)Jin, Yu, Huo, Wang, Wang, He, and Han]{ugcn}
Jin, D., Yu, Z., Huo, C., Wang, R., Wang, X., He, D., and Han, J.
\newblock Universal graph convolutional networks.
\newblock \emph{Advances in Neural Information Processing Systems},
  34:\penalty0 10654--10664, 2021.

\bibitem[Kang et~al.(2019)Kang, Xie, Rohrbach, Yan, Gordo, Feng, and
  Kalantidis]{decoupling}
Kang, B., Xie, S., Rohrbach, M., Yan, Z., Gordo, A., Feng, J., and Kalantidis,
  Y.
\newblock Decoupling representation and classifier for long-tailed recognition.
\newblock In \emph{International Conference on Learning Representations}, 2019.

\bibitem[Kingma \& Ba(2015)Kingma and Ba]{adam}
Kingma, D.~P. and Ba, J.
\newblock Adam: {A} method for stochastic optimization.
\newblock In \emph{3rd International Conference on Learning Representations,
  {ICLR} 2015}, 2015.

\bibitem[Lim et~al.(2021)Lim, Hohne, Li, Huang, Gupta, Bhalerao, and
  Lim]{classhomo}
Lim, D., Hohne, F., Li, X., Huang, S.~L., Gupta, V., Bhalerao, O., and Lim,
  S.~N.
\newblock Large scale learning on non-homophilous graphs: New benchmarks and
  strong simple methods.
\newblock \emph{Advances in Neural Information Processing Systems},
  34:\penalty0 20887--20902, 2021.

\bibitem[Lin et~al.(2017)Lin, Goyal, Girshick, He, and Doll{\'a}r]{focal}
Lin, T.-Y., Goyal, P., Girshick, R., He, K., and Doll{\'a}r, P.
\newblock Focal loss for dense object detection.
\newblock In \emph{Proceedings of the IEEE international conference on computer
  vision}, pp.\  2980--2988, 2017.

\bibitem[Liu et~al.(2021)Liu, Ao, Qin, Chi, Feng, Yang, and He]{pc_gnn}
Liu, Y., Ao, X., Qin, Z., Chi, J., Feng, J., Yang, H., and He, Q.
\newblock Pick and choose: a gnn-based imbalanced learning approach for fraud
  detection.
\newblock In \emph{Proceedings of the Web Conference 2021}, pp.\  3168--3177,
  2021.

\bibitem[Menon et~al.(2020)Menon, Jayasumana, Rawat, Jain, Veit, and
  Kumar]{logit_adjustment}
Menon, A.~K., Jayasumana, S., Rawat, A.~S., Jain, H., Veit, A., and Kumar, S.
\newblock Long-tail learning via logit adjustment.
\newblock In \emph{International Conference on Learning Representations}, 2020.

\bibitem[Mohammadrezaei et~al.(2018)Mohammadrezaei, Shiri, and
  Rahmani]{DBLP:journals/scn/MohammadrezaeiS18}
Mohammadrezaei, M., Shiri, M.~E., and Rahmani, A.~M.
\newblock Identifying fake accounts on social networks based on graph analysis
  and classification algorithms.
\newblock \emph{Secur. Commun. Networks}, 2018:\penalty0 5923156:1--5923156:8,
  2018.

\bibitem[Park et~al.(2021)Park, Song, and Yang]{graphens}
Park, J., Song, J., and Yang, E.
\newblock Graphens: Neighbor-aware ego network synthesis for class-imbalanced
  node classification.
\newblock In \emph{International Conference on Learning Representations}, 2021.

\bibitem[Pei et~al.(2019)Pei, Wei, Chang, Lei, and Yang]{geomgcn}
Pei, H., Wei, B., Chang, K. C.-C., Lei, Y., and Yang, B.
\newblock Geom-gcn: Geometric graph convolutional networks.
\newblock In \emph{International Conference on Learning Representations}, 2019.

\bibitem[Qu et~al.(2021)Qu, Zhu, Zheng, Shi, and Yin]{imgagn}
Qu, L., Zhu, H., Zheng, R., Shi, Y., and Yin, H.
\newblock Imgagn: Imbalanced network embedding via generative adversarial graph
  networks.
\newblock In \emph{Proceedings of the 27th ACM SIGKDD Conference on Knowledge
  Discovery \& Data Mining}, pp.\  1390--1398, 2021.

\bibitem[Ren et~al.(2020)Ren, Yu, Ma, Zhao, Yi, et~al.]{balanced_softmax}
Ren, J., Yu, C., Ma, X., Zhao, H., Yi, S., et~al.
\newblock Balanced meta-softmax for long-tailed visual recognition.
\newblock \emph{Advances in neural information processing systems},
  33:\penalty0 4175--4186, 2020.

\bibitem[Rozemberczki et~al.(2021)Rozemberczki, Allen, and Sarkar]{wikipedia}
Rozemberczki, B., Allen, C., and Sarkar, R.
\newblock Multi-scale attributed node embedding.
\newblock \emph{J. Complex Networks}, 9\penalty0 (2), 2021.

\bibitem[Sen et~al.(2008)Sen, Namata, Bilgic, Getoor, Gallagher, and
  Eliassi{-}Rad]{DBLP:journals/aim/SenNBGGE08}
Sen, P., Namata, G., Bilgic, M., Getoor, L., Gallagher, B., and Eliassi{-}Rad,
  T.
\newblock Collective classification in network data.
\newblock \emph{{AI} Mag.}, 29\penalty0 (3):\penalty0 93--106, 2008.

\bibitem[Shi et~al.(2020)Shi, Tang, Zhu, Wilson, and Liu]{conditionalgan}
Shi, M., Tang, Y., Zhu, X., Wilson, D., and Liu, J.
\newblock Multi-class imbalanced graph convolutional network learning.
\newblock In \emph{Proceedings of the Twenty-Ninth International Joint
  Conference on Artificial Intelligence (IJCAI-20)}, 2020.

\bibitem[Srivastava et~al.(2014)Srivastava, Hinton, Krizhevsky, Sutskever, and
  Salakhutdinov]{dropout}
Srivastava, N., Hinton, G.~E., Krizhevsky, A., Sutskever, I., and
  Salakhutdinov, R.
\newblock Dropout: a simple way to prevent neural networks from overfitting.
\newblock \emph{J. Mach. Learn. Res.}, 15\penalty0 (1):\penalty0 1929--1958,
  2014.

\bibitem[Tan et~al.(2020)Tan, Wang, Li, Li, Ouyang, Yin, and Yan]{equalization}
Tan, J., Wang, C., Li, B., Li, Q., Ouyang, W., Yin, C., and Yan, J.
\newblock Equalization loss for long-tailed object recognition.
\newblock In \emph{Proceedings of the IEEE/CVF conference on computer vision
  and pattern recognition}, pp.\  11662--11671, 2020.

\bibitem[Tang et~al.(2020)Tang, Huang, and Zhang]{causal_norm}
Tang, K., Huang, J., and Zhang, H.
\newblock Long-tailed classification by keeping the good and removing the bad
  momentum causal effect.
\newblock \emph{Advances in Neural Information Processing Systems},
  33:\penalty0 1513--1524, 2020.

\bibitem[Tian et~al.(2020)Tian, Liu, Glaser, Hsu, and Kira]{post_ic}
Tian, J., Liu, Y.-C., Glaser, N., Hsu, Y.-C., and Kira, Z.
\newblock Posterior re-calibration for imbalanced datasets.
\newblock \emph{Advances in Neural Information Processing Systems},
  33:\penalty0 8101--8113, 2020.

\bibitem[Veli{\v{c}}kovi{\'c} et~al.(2018)Veli{\v{c}}kovi{\'c}, Cucurull,
  Casanova, Romero, Li{\`o}, and Bengio]{gat}
Veli{\v{c}}kovi{\'c}, P., Cucurull, G., Casanova, A., Romero, A., Li{\`o}, P.,
  and Bengio, Y.
\newblock Graph attention networks.
\newblock In \emph{International Conference on Learning Representations}, 2018.

\bibitem[Wang et~al.(2021{\natexlab{a}})Wang, Lukasiewicz, Hu, Cai, and
  Xu]{rsg}
Wang, J., Lukasiewicz, T., Hu, X., Cai, J., and Xu, Z.
\newblock Rsg: A simple but effective module for learning imbalanced datasets.
\newblock In \emph{Proceedings of the IEEE/CVF Conference on Computer Vision
  and Pattern Recognition}, pp.\  3784--3793, 2021{\natexlab{a}}.

\bibitem[Wang et~al.(2021{\natexlab{b}})Wang, Zhang, Zang, Cao, Pang, Gong,
  Chen, Liu, Loy, and Lin]{seesaw}
Wang, J., Zhang, W., Zang, Y., Cao, Y., Pang, J., Gong, T., Chen, K., Liu, Z.,
  Loy, C.~C., and Lin, D.
\newblock Seesaw loss for long-tailed instance segmentation.
\newblock In \emph{Proceedings of the IEEE/CVF conference on computer vision
  and pattern recognition}, pp.\  9695--9704, 2021{\natexlab{b}}.

\bibitem[Wang et~al.(2021{\natexlab{c}})Wang, Zhu, Zhao, Zeng, Wang, and
  Tang]{acl}
Wang, T., Zhu, Y., Zhao, C., Zeng, W., Wang, J., and Tang, M.
\newblock Adaptive class suppression loss for long-tail object detection.
\newblock In \emph{Proceedings of the IEEE/CVF conference on computer vision
  and pattern recognition}, pp.\  3103--3112, 2021{\natexlab{c}}.

\bibitem[Wang et~al.(2020{\natexlab{a}})Wang, Lian, Miao, Liu, and Yu]{ride}
Wang, X., Lian, L., Miao, Z., Liu, Z., and Yu, S.
\newblock Long-tailed recognition by routing diverse distribution-aware
  experts.
\newblock In \emph{International Conference on Learning Representations},
  2020{\natexlab{a}}.

\bibitem[Wang et~al.(2020{\natexlab{b}})Wang, Ye, Wang, Cui, and Yu]{rect}
Wang, Z., Ye, X., Wang, C., Cui, J., and Yu, P.
\newblock Network embedding with completely-imbalanced labels.
\newblock \emph{IEEE Transactions on Knowledge and Data Engineering},
  2020{\natexlab{b}}.

\bibitem[Welling \& Kipf(2016)Welling and Kipf]{gcn}
Welling, M. and Kipf, T.~N.
\newblock Semi-supervised classification with graph convolutional networks.
\newblock In \emph{J. International Conference on Learning Representations
  (ICLR 2017)}, 2016.

\bibitem[Xiang et~al.(2020)Xiang, Ding, and Han]{lmfe}
Xiang, L., Ding, G., and Han, J.
\newblock Learning from multiple experts: Self-paced knowledge distillation for
  long-tailed classification.
\newblock In \emph{European Conference on Computer Vision}, pp.\  247--263.
  Springer, 2020.

\bibitem[Xu et~al.(2020)Xu, Dan, Khim, and Ravikumar]{lcbal}
Xu, Z., Dan, C., Khim, J., and Ravikumar, P.
\newblock Class-weighted classification: Trade-offs and robust approaches.
\newblock In \emph{International Conference on Machine Learning}, pp.\
  10544--10554. PMLR, 2020.

\bibitem[Yang et~al.(2021)Yang, Li, Liu, Wang, Cao, Guo, et~al.]{dmp}
Yang, L., Li, M., Liu, L., Wang, C., Cao, X., Guo, Y., et~al.
\newblock Diverse message passing for attribute with heterophily.
\newblock \emph{Advances in Neural Information Processing Systems},
  34:\penalty0 4751--4763, 2021.

\bibitem[Yang et~al.(2016)Yang, Cohen, and Salakhudinov]{public_split}
Yang, Z., Cohen, W., and Salakhudinov, R.
\newblock Revisiting semi-supervised learning with graph embeddings.
\newblock In \emph{International conference on machine learning}, pp.\  40--48.
  PMLR, 2016.

\bibitem[Ying et~al.(2018)Ying, He, Chen, Eksombatchai, Hamilton, and
  Leskovec]{DBLP:conf/kdd/YingHCEHL18}
Ying, R., He, R., Chen, K., Eksombatchai, P., Hamilton, W.~L., and Leskovec, J.
\newblock Graph convolutional neural networks for web-scale recommender
  systems.
\newblock In \emph{Proceedings of the 24th ACM SIGKDD international conference
  on knowledge discovery \& data mining}, pp.\  974--983, 2018.

\bibitem[Zhang et~al.(2021)Zhang, Wei, Zhou, and Wu]{bag_tricks}
Zhang, Y., Wei, X.-S., Zhou, B., and Wu, J.
\newblock Bag of tricks for long-tailed visual recognition with deep
  convolutional neural networks.
\newblock In \emph{Proceedings of the AAAI conference on artificial
  intelligence}, volume~35, pp.\  3447--3455, 2021.

\bibitem[Zhao et~al.(2021)Zhao, Zhang, and Wang]{graphsmote}
Zhao, T., Zhang, X., and Wang, S.
\newblock Graphsmote: Imbalanced node classification on graphs with graph
  neural networks.
\newblock In \emph{Proceedings of the 14th ACM international conference on web
  search and data mining}, pp.\  833--841, 2021.

\bibitem[Zhou et~al.(2020)Zhou, Cui, Wei, and Chen]{bbn}
Zhou, B., Cui, Q., Wei, X.-S., and Chen, Z.-M.
\newblock Bbn: Bilateral-branch network with cumulative learning for
  long-tailed visual recognition.
\newblock In \emph{Proceedings of the IEEE/CVF conference on computer vision
  and pattern recognition}, pp.\  9719--9728, 2020.

\bibitem[Zhu et~al.(2020)Zhu, Yan, Zhao, Heimann, Akoglu, and Koutra]{h2gnn}
Zhu, J., Yan, Y., Zhao, L., Heimann, M., Akoglu, L., and Koutra, D.
\newblock Beyond homophily in graph neural networks: Current limitations and
  effective designs.
\newblock \emph{Advances in Neural Information Processing Systems},
  33:\penalty0 7793--7804, 2020.

\bibitem[Zhu et~al.(2021)Zhu, Rossi, Rao, Mai, Lipka, Ahmed, and Koutra]{cpgnn}
Zhu, J., Rossi, R.~A., Rao, A., Mai, T., Lipka, N., Ahmed, N.~K., and Koutra,
  D.
\newblock Graph neural networks with heterophily.
\newblock In \emph{Proceedings of the AAAI Conference on Artificial
  Intelligence}, volume~35, pp.\  11168--11176, 2021.

\end{thebibliography}
\bibliographystyle{icml2022}

\newpage
\appendix
\onecolumn
\section{Additional Experimental Results} \label{appen:results}
In this section, we provide additional experimental results which are omitted due to the space constraints.

\subsection{The False Negative Rates in Section~\ref{sec:problem}} \label{appensub:problem}
To prove that TAM does not significantly sacrifice the false negative rates for reducing false positive cases, we present the false negative rates of the results in Section~\ref{sec:problem}. As shown in Table~\ref{tb:problem_all}, our approach effectively reduces the false positive cases for major nodes connected with non-typical minor nodes while increasing false negative rates slightly. These consistent results over multiple benchmark datasets strengthen our claim that TAM can successfully mitigate the false positive cases of major classes.


\begin{table}[h]
\caption{\small Comparison of false positive rates (FPR) and false negative rates (FNR) near normal minor nodes and anomalously-connected minor nodes.}
\begin{center}
\begin{small}
\setlength{\tabcolsep}{1.30pt} 
\renewcommand{\arraystretch}{1} 
\begin{tabular}{@{\extracolsep{1pt}}lcc|cc|cc|cc@{}}
\toprule
\multirow{2}{*}{\shortstack[1]{\textbf{Method} \\ (GCN)}}& \multicolumn{2}{c}{Cora\tiny{$\rho$=\textbf{10}}} & \multicolumn{2}{c}{CiteSeer\tiny{$\rho$=\textbf{10}}} & \multicolumn{2}{c}{Chameleon\tiny{$\rho$=\textbf{5}}} & \multicolumn{2}{c}{Wisconsin\tiny{$\rho$=\textbf{12}}} \\
\cline{2-9}
  & FPR & FNR & FPR & FNR & FPR & FNR & FPR & FNR \\
\cline{1-9}
                     Re-Weight & 47.68 & 9.97 & 64.63 & 4.74 & 48.57 & 29.06 & 63.90 & 36.40 \\
                     + \textbf{TAM} & 29.77\scriptsize{\textbf{(-17.91)}} & 15.83\scriptsize{\textbf{(+5.86)}}
                     & 57.84\scriptsize{\textbf{(-6.79)}} & 4.78\scriptsize{\textbf{(+0.04)}} 
                     & 35.78\scriptsize{\textbf{(-12.79)}} & 33.73\scriptsize{\textbf{(+4.67)}}
                     & 34.63\scriptsize{\textbf{(-29.27)}} & 39.67\scriptsize{\textbf{(+3.27)}} \\
                     \cdashline{1-9}
                     BalancedSoftmax & 56.69 & 5.91 & 65.86 & 4.95 & 41.85 & 31.37 & 52.19 & 45.77 \\
                     + \textbf{TAM} & 37.97\scriptsize{\textbf{(-18.72)}} & 11.78\scriptsize{\textbf{(+5.87)}} & 56.32\scriptsize{\textbf{(-9.54)}} & 5.85\scriptsize{\textbf{(+0.90)}} & 36.01\scriptsize{\textbf{(-5.84)}} & 28.27\scriptsize{\textbf{(-3.10)}} & 40.14\scriptsize{\textbf{(-12.05)}} & 40.46\scriptsize{\textbf{(-5.31)}} \\
                     \cdashline{1-9}
                     GraphSMOTE  & 45.19 & 10.51 & 62.55 & 3.80 & 49.81 & 19.48 & 52.90 & 47.33 \\
                     + \textbf{TAM} & 29.82\scriptsize{\textbf{(-15.37)}} & 15.58\scriptsize{\textbf{(+5.07)}} 
                     & 54.14\scriptsize{\textbf{(-8.41)}} & 5.59\scriptsize{\textbf{(+1.79)}} 
                     & 48.73\scriptsize{\textbf{(-1.08)}} & 16.45\scriptsize{\textbf{(-3.03)}}
                     & 48.70\scriptsize{\textbf{(-4.20)}} & 50.30\scriptsize{\textbf{(+2.97)}} \\
\bottomrule
\end{tabular}
\end{small}
\end{center}
\label{tb:problem_all}
\end{table}

\subsection{The Results of Three Benchmark Datasets (Homophilous Graphs)} \label{appensub:homo}
In the main paper, we only report the comparison of our method with other baselines on Cora, CiteSeer, and PubMed~\citep{DBLP:journals/aim/SenNBGGE08} with the high imbalance ratio ($\rho=10$) due to the space limitation. To show that our method is also effective under low imbalance ratio, we provide the results on Cora, CiteSeer, and PubMed with relatively low imbalance ratio ($\rho=5$) over three GNN architectures: GCN, GAT, and SAGE in Table~\ref{tb:appx_main_homo}. We observe consistent results with the main paper in that our method improves various imbalance handling algorithms by combining ours with these baselines.

\begin{table*}[h]
\caption{\small Experimental results of our algorithm TAM and other baselines on three class-imbalanced node classification benchmark datasets (homophilous graphs). We report averaged balanced accuracy (bAcc.) and F1-score with the standard errors for 10 repetitions on three representative GNN architectures.}
\begin{center}
\begin{footnotesize}
\setlength{\columnsep}{1pt}%
\begin{adjustbox}{width=0.90\linewidth}
\begin{tabular}{@{\extracolsep{1pt}}rlcc|cc|cc@{}}
\toprule
 & \multirow{1}{*}{\textbf{Dataset}} & \multicolumn{2}{c}{Cora} & \multicolumn{2}{c}{CiteSeer} & \multicolumn{2}{c}{PubMed}  \\ 
\cline{2-8}\rule{0pt}{2.2ex}
& \textbf{Imbalance Ratio} ($\rho=5$) & bAcc. & F1 & bAcc. & F1 & bAcc. & F1 \\
\cline{2-8}
\rule{0pt}{2.5ex}  
\multirow{11}{*}{\rotatebox{90}{GCN}} & Cross Entropy & 69.15 \tiny{$\pm 0.52$} & 69.36 \tiny{$\pm 0.75$}
                   
                    & 48.56 \tiny{$\pm 1.70$}& 44.56 \tiny{$\pm 2.30$} 
                    
                    & 71.89 \tiny{$\pm 1.04$}& 67.59 \tiny{$\pm 1.54$}
                    
                    \\
                    \rule{0pt}{2ex}
                     & Re-Weight & 71.86 \tiny{$\pm 0.76$} & 72.23 \tiny{$\pm 0.75$} 
                     
                     & 54.32 \tiny{$\pm 1.33$}& 52.37 \tiny{$\pm 1.65$} 
                     
                     & 73.91 \tiny{$\pm 0.95$}& 71.70 \tiny{$\pm 0.77$}
                     
                     \\
                    
                     & PC Softmax & 72.69 \tiny{$\pm 0.58$} & 72.90 \tiny{$\pm 0.56$}
                    
                     & 58.86 \tiny{$\pm 1.27$}& 57.33 \tiny{$\pm 1.46$}
                    
                     & 74.13 \tiny{$\pm 0.73$}& 72.84 \tiny{$\pm 0.72$}
                   
                     \\
                     
                     & DR-GCN & 67.56 \tiny{$\pm 0.56$} & 67.29 \tiny{$\pm 0.73$} 
                     
                     & 50.47 \tiny{$\pm 1.17$}& 47.73 \tiny{$\pm 1.59$}
                    
                     & 70.36 \tiny{$\pm 0.66$}& 68.22 \tiny{$\pm 0.93$}
                     
                     \\

                     & GraphSMOTE & 73.46 \tiny{$\pm 0.84$} & 73.07 \tiny{$\pm 0.67$}
                     
                     & 54.79 \tiny{$\pm 1.21$}& 53.52 \tiny{$\pm 1.46$}  
                     
                     & 72.49 \tiny{$\pm 0.79$}& 69.80 \tiny{$\pm 1.17$}
                    
                     \\
                     
                     \cline{2-8}
                     
                     & BalancedSoftmax & 73.63 \tiny{$\pm 0.68$} & 73.40 \tiny{$\pm 0.67$}
                    
                     & 60.13 \tiny{$\pm 1.48$}& 59.30 \tiny{$\pm 1.61$}
                   
                     & 75.26 \tiny{$\pm 0.58$}& 73.83 \tiny{$\pm 0.66$}
                  
                     \\
                     & + \textbf{TAM} & 73.75 \tiny{$\pm 0.66$} & 73.74 \tiny{$\pm 0.66$}
                     
                     & 60.97 \tiny{$\pm 1.02$}& 60.46 \tiny{$\pm 0.98$}
                   
                     & \textbf{76.03} \tiny{$\pm 0.96$}& 75.16 \tiny{$\pm 1.10$}
                    
                     \\
                     \cdashline{2-8}
                     & ReNode & 74.91 \tiny{$\pm 0.57$} & 75.37 \tiny{$\pm 0.62$} 
                     
                     & 58.01 \tiny{$\pm 1.52$}& 56.63 \tiny{$\pm 1.85$}
                    
                     & 73.99 \tiny{$\pm 0.88$}& 72.04 \tiny{$\pm 1.01$}
                    
                     \\
                     & + \textbf{TAM} & 74.77 \tiny{$\pm 0.42$} & 75.57 \tiny{$\pm 0.43$} 
                   
                     & 58.57 \tiny{$\pm 1.59$}& 57.56 \tiny{$\pm 1.80$}
                     
                     & 75.22 \tiny{$\pm 1.05$}& 74.22 \tiny{$\pm 1.41$}
                    
                     \\
                     \cdashline{2-8}
                     & GraphENS & 75.68 \tiny{$\pm 0.58$} & 75.47 \tiny{$\pm 0.58$}
                     
                     & 62.24 \tiny{$\pm 1.10$}& 61.70 \tiny{$\pm 1.11$}
                    
                     & 74.30 \tiny{$\pm 0.59$}& 73.53 \tiny{$\pm 0.52$}
                     
                     \\
                     & + \textbf{TAM} & \textbf{75.72} \tiny{$\pm 0.64$} & \textbf{75.93} \tiny{$\pm 0.62$} 
                    
                     & \textbf{63.01} \tiny{$\pm 0.87$}& \textbf{62.56} \tiny{$\pm 0.80$}
                    
                     & 75.62 \tiny{$\pm 0.55$}& \textbf{75.28} \tiny{$\pm 0.51$}
                    
                     \\
\cline{2-8}
\noalign{\vskip\doublerulesep
         \vskip-\arrayrulewidth} \cline{2-8}
\rule{0pt}{2.5ex}  
\multirow{11}{*}{\rotatebox{90}{GAT}} & Cross Entropy & 68.12 \tiny{$\pm 0.51$}& 68.81 \tiny{$\pm 0.62$}
                   
                    & 51.43 \tiny{$\pm 1.67$}& 48.85 \tiny{$\pm 2.13$}
                   
                    & 70.65 \tiny{$\pm 1.11$}& 66.73 \tiny{$\pm 1.69$}
                    
                    \\
                    \rule{0pt}{2ex}
                     & Re-Weight & 73.24 \tiny{$\pm 0.81$}& 72.40 \tiny{$\pm 0.96$}
                   
                     & 55.40 \tiny{$\pm 1.59$}& 53.97 \tiny{$\pm 1.62$}
                   
                     & 72.94 \tiny{$\pm 0.77$}& 70.59 \tiny{$\pm 1.10$}
                    
                     \\
                    
                     & PC Softmax & 71.24 \tiny{$\pm 0.52$}& 71.53 \tiny{$\pm 0.62$}
                   
                     & 58.83 \tiny{$\pm 1.28$}& 57.45 \tiny{$\pm 1.37$}
                   
                     & 74.72 \tiny{$\pm 0.69$}& 72.66 \tiny{$\pm 0.82$}
                   
                     \\
                     
                     & DR-GCN & 66.43 \tiny{$\pm 0.72$}& 66.31 \tiny{$\pm 0.84$}
                    
                     & 51.48 \tiny{$\pm 1.63$}& 49.48 \tiny{$\pm 2.31$}
                    
                     & 72.41 \tiny{$\pm 0.57$}& 71.74 \tiny{$\pm 0.63$}
                   
                     \\

                     & GraphSMOTE & 72.96 \tiny{$\pm 0.67$}& 72.39 \tiny{$\pm 0.83$}
                    
                     & 55.38 \tiny{$\pm 1.52$}& 53.72 \tiny{$\pm 1.88$}
                    
                     & 72.94 \tiny{$\pm 0.85$}& 70.65 \tiny{$\pm 1.23$}
                   
                     \\
                     
                     \cline{2-8}
                     
                     & BalancedSoftmax & 72.50 \tiny{$\pm 0.60$}& 71.96 \tiny{$\pm 0.67$}
                     
                     & 59.72 \tiny{$\pm 1.15$}& 58.79 \tiny{$\pm 1.18$}
                     
                     & 73.38 \tiny{$\pm 0.74$}& 72.47 \tiny{$\pm 0.83$}
                
                     \\
                     & + \textbf{TAM} & 72.72 \tiny{$\pm 0.66$}& 72.78 \tiny{$\pm 0.81$}
                     
                     & 62.19 \tiny{$\pm 0.87$}& 61.55 \tiny{$\pm 0.86$}
                    
                     & 74.71 \tiny{$\pm 0.74$}& 74.14 \tiny{$\pm 0.80$}
                   
                     \\
                     \cdashline{2-8}
                     & ReNode & 74.34 \tiny{$\pm 0.69$}& 74.77 \tiny{$\pm 0.52$}
                    
                     & 58.69 \tiny{$\pm 1.64$}& 57.05 \tiny{$\pm 1.94$}
                    
                     & 73.85 \tiny{$\pm 0.96$}& 71.79 \tiny{$\pm 1.16$}
                  
                     \\
                     & + \textbf{TAM} & \textbf{75.07} \tiny{$\pm 0.62$}& 75.05 \tiny{$\pm 0.69$}
                   
                     & 59.11 \tiny{$\pm 1.41$}& 57.77 \tiny{$\pm 1.55$}
                   
                     & 73.79 \tiny{$\pm 0.91$}& 72.30 \tiny{$\pm 0.88$}
                     
                     \\
                     \cdashline{2-8}
                     & GraphENS & 74.92 \tiny{$\pm 0.57$}& 74.58 \tiny{$\pm 0.61$}
                    
                     & 59.40 \tiny{$\pm 1.08$}& 58.98 \tiny{$\pm 1.11$}
                    
                     & 73.93 \tiny{$\pm 0.66$}& 72.99 \tiny{$\pm 0.90$}
                    
                     \\
                     & + \textbf{TAM} & 74.82 \tiny{$\pm 0.40$}& \textbf{75.13} \tiny{$\pm 0.43$}
                     
                     & \textbf{62.23} \tiny{$\pm 0.79$}& \textbf{61.89} \tiny{$\pm 0.79$}
                    
                     & \textbf{75.05} \tiny{$\pm 0.65$}& \textbf{74.47} \tiny{$\pm 0.66$}
                     
                     \\
\cline{2-8}
\noalign{\vskip\doublerulesep
         \vskip-\arrayrulewidth} \cline{2-8}
\rule{0pt}{2.5ex}  
\multirow{11}{*}{\rotatebox{90}{SAGE}} & Cross Entropy & 66.58 \tiny{$\pm 0.78$}& 66.43 \tiny{$\pm 0.86$}
                    
                    & 51.50 \tiny{$\pm 1.55$}& 49.01 \tiny{$\pm 2.09$}
                    
                    & 71.55 \tiny{$\pm 0.74$}& 70.38 \tiny{$\pm 0.73$}
                   
                    \\
                    \rule{0pt}{2ex}
                     & Re-Weight & 71.59 \tiny{$\pm 0.77$}& 71.91 \tiny{$\pm 0.87$}
                     
                     & 56.65 \tiny{$\pm 1.50$}& 55.38 \tiny{$\pm 1.72$}
                     
                     & 72.22 \tiny{$\pm 0.95$}& 70.33 \tiny{$\pm 0.99$}
                   
                     \\
                    
                     & PC Softmax & 71.55 \tiny{$\pm 0.72$}& 71.22 \tiny{$\pm 0.80$}
                   
                     & 56.85 \tiny{$\pm 1.52$}& 55.27 \tiny{$\pm 1.73$}
                   
                     & 73.21 \tiny{$\pm 0.46$}& 72.33 \tiny{$\pm 0.63$}
                    
                     \\
                     
                     & DR-GCN & 66.20 \tiny{$\pm 0.68$}& 66.03 \tiny{$\pm 0.73$}
                     
                     & 54.31 \tiny{$\pm 1.42$}& 53.36 \tiny{$\pm 1.44$}
                    
                     & 71.43 \tiny{$\pm 0.83$}& 70.22 \tiny{$\pm 1.03$}
                    
                     \\

                     & GraphSMOTE & 69.66 \tiny{$\pm 0.78$}& 69.98 \tiny{$\pm 0.89$}
                    
                     & 52.90 \tiny{$\pm 1.19$}& 50.70 \tiny{$\pm 1.72$}  
                    
                     & 70.71 \tiny{$\pm 1.39$}& 69.12 \tiny{$\pm 1.68$}
                    
                     \\
                     
                     \cline{2-8}
                     
                     & BalancedSoftmax & 71.50 \tiny{$\pm 0.43$}& 71.68 \tiny{$\pm 0.47$}
                    
                     & 58.49 \tiny{$\pm 1.32$}& 57.91 \tiny{$\pm 1.44$}
                    
                     & 72.82 \tiny{$\pm 0.53$}& 71.64 \tiny{$\pm 0.62$}
                    
                     \\
                     & + \textbf{TAM} & 72.86 \tiny{$\pm 0.39$}& 72.81 \tiny{$\pm 0.44$}
                    
                     & 61.09 \tiny{$\pm 1.28$}& 60.41 \tiny{$\pm 1.22$}
                    
                     & \textbf{74.37} \tiny{$\pm 0.56$}& \textbf{73.98} \tiny{$\pm 0.61$}
                    
                     \\
                     \cdashline{2-8}
                     & ReNode & 72.92 \tiny{$\pm 0.48$}& 73.58 \tiny{$\pm 0.50$}
                     
                     & 58.36 \tiny{$\pm 1.69$}& 57.09 \tiny{$\pm 2.08$}
                     
                     & 73.51 \tiny{$\pm 1.04$}& 71.98 \tiny{$\pm 1.07$}
                     
                     \\
                     & + \textbf{TAM} & 73.09 \tiny{$\pm 0.34$}& 73.76 \tiny{$\pm 0.35$}
                     
                     & 58.69 \tiny{$\pm 1.28$}& 57.50 \tiny{$\pm 1.51$}
                    
                     & 73.84 \tiny{$\pm 0.64$}& 72.86 \tiny{$\pm 0.92$}
                     
                     \\
                     \cdashline{2-8}
                     & GraphENS & 73.43 \tiny{$\pm 0.62$}& 73.47 \tiny{$\pm 0.74$}
                    
                     & 60.17 \tiny{$\pm 1.33$}& 59.71 \tiny{$\pm 1.32$}
                   
                     & 73.32 \tiny{$\pm 0.75$}& 72.74 \tiny{$\pm 0.68$}
                 
                     \\
                     & + \textbf{TAM} & \textbf{75.27} \tiny{$\pm 0.37$}& \textbf{75.36} \tiny{$\pm 0.59$}
                   
                     & \textbf{62.40} \tiny{$\pm 1.10$}& \textbf{61.96} \tiny{$\pm 1.05$}
                     
                     & 73.93 \tiny{$\pm 0.60$}& 73.74 \tiny{$\pm 0.47$}
                
                     \\

\bottomrule

\end{tabular}
\end{adjustbox}
\end{footnotesize}
\end{center}
\label{tb:appx_main_homo}
\end{table*}

\subsection{The Results of Two Benchmark Datasets (Heterophilous Graphs)} \label{appensub:hetero}

We also provide the additional experimental results on heterophilous graphs: Chameleon and Squirrel~\citep{wikipedia} with the different imbalance ratio ($\rho=10$). In Table~\ref{tb:appx_main_hetero}, we affirm consistent results with the main paper in that our method improves various imbalance handling algorithms by combining ours with these baselines in most cases. For the Squirrel dataset, our method shows comparable performance with baselines. We conjecture that low accuracy of baselines in Squirrel induces erroneous estimation of neighbor label distribution and class-wise connectivity matrix, resulting in insignificant improvements. 

\begin{table*}[t]
\caption{\small Experimental results of our algorithm TAM and other baselines on two class-imbalanced node classification benchmark datasets (heterophilous graphs). We report averaged balanced accuracy (bAcc.) and F1-score with the standard errors for 10 repetitions on three representative GNN architectures.}
\begin{center}
\begin{footnotesize}
\setlength{\columnsep}{1pt}%
\begin{adjustbox}{width=0.73\linewidth}
\begin{tabular}{@{\extracolsep{1pt}}rlcc|cc@{}}
\toprule
 & \multirow{1}{*}{\textbf{Dataset}} & \multicolumn{2}{c}{Chameleon} & \multicolumn{2}{c}{Squirrel}  \\ 
\cline{2-6}\rule{0pt}{2.2ex}
& \textbf{Imbalance Ratio} ($\rho=10$) & bAcc. & F1 & bAcc. & F1 \\
\cline{2-6}
\rule{0pt}{2.5ex}  
\multirow{11}{*}{\rotatebox{90}{GCN}} & Cross Entropy 
                    & 31.52 \tiny{$\pm 0.72$}& 29.30 \tiny{$\pm 0.73$}
                    
                    & 24.76 \tiny{$\pm 0.37$}& 18.57 \tiny{$\pm 0.27$}
                   
                    \\
                    \rule{0pt}{2ex}
                     & Re-Weight 
                     & 36.07 \tiny{$\pm 0.87$}& 35.61 \tiny{$\pm 0.81$}
                    
                     & 26.92 \tiny{$\pm 0.53$}& 25.04 \tiny{$\pm 0.59$}
                    
                     \\
                    
                     & PC Softmax
                     & 36.86 \tiny{$\pm 1.04$}& 36.24 \tiny{$\pm 1.01$}
                    
                     & 26.49 \tiny{$\pm 0.59$}& 25.73 \tiny{$\pm 0.49$}
                     
                     \\
                     
                     & DR-GCN 
                     & 33.34 \tiny{$\pm 0.81$}& 29.60 \tiny{$\pm 0.79$}
                     
                     & 23.34 \tiny{$\pm 0.43$}& 18.20 \tiny{$\pm 0.49$}
                     
                     \\

                     & GraphENS
                     & 41.54 \tiny{$\pm 0.63$}& 40.19 \tiny{$\pm 0.68$}
                    
                     & 26.75 \tiny{$\pm 0.35$}& 26.31 \tiny{$\pm 0.27$}
                    
                     \\
                     
                     \cline{2-6}
                     
                     & BalancedSoftmax 
                     & 36.47 \tiny{$\pm 0.89$}& 35.94 \tiny{$\pm 0.85$}
                     
                     & 27.32 \tiny{$\pm 0.52$}& 26.42 \tiny{$\pm 0.41$}
                     \\
                     & + \textbf{TAM} 
                     & 38.85 \tiny{$\pm 1.00$}& 37.44 \tiny{$\pm 0.96$}
                    
                     & 27.81 \tiny{$\pm 0.45$}& 27.25 \tiny{$\pm 0.45$}
                    
                     \\
                     \cdashline{2-6}
                     & ReNode 
                     & 34.26 \tiny{$\pm 1.13$}& 33.66 \tiny{$\pm 1.09$}
                   
                     & 25.42 \tiny{$\pm 0.34$}& 24.55 \tiny{$\pm 0.41$}
                    
                     \\
                     & + \textbf{TAM} 
                     & 38.01 \tiny{$\pm 0.97$}& 36.92 \tiny{$\pm 0.94$}
                    
                     & 26.41 \tiny{$\pm 0.36$}& 25.87 \tiny{$\pm 0.43$}
                   
                     \\
                     \cdashline{2-6}
                     & GraphSMOTE 
                     & 41.50 \tiny{$\pm 0.82$}& 40.80 \tiny{$\pm 0.79$}
                     
                     & 27.14 \tiny{$\pm 0.49$}& 26.67 \tiny{$\pm 0.53$}
                   
                     \\
                     & + \textbf{TAM} 
                     & \textbf{42.80} \tiny{$\pm 0.89$}& \textbf{41.91} \tiny{$\pm 0.88$}
                    
                     & \textbf{28.30} \tiny{$\pm 0.46$}& \textbf{27.81} \tiny{$\pm 0.48$}
                    
                     \\
\cline{2-6}
\noalign{\vskip\doublerulesep
         \vskip-\arrayrulewidth} \cline{2-6}
\rule{0pt}{2.5ex}  
\multirow{11}{*}{\rotatebox{90}{GAT}} & Cross Entropy 
                    & 32.41 \tiny{$\pm 0.70$}& 27.33 \tiny{$\pm 0.94$}
                   
                    & 24.69 \tiny{$\pm 0.39$}& 18.89 \tiny{$\pm 0.38$}
                  
                    \\
                    \rule{0pt}{2ex}
                     & Re-Weight 
                     & 35.72 \tiny{$\pm 0.65$}& 34.19 \tiny{$\pm 0.74$}
                    
                     & 25.79 \tiny{$\pm 0.52$}& 24.32 \tiny{$\pm 0.62$}
                 
                     \\
                    
                     & PC Softmax
                     & 38.32 \tiny{$\pm 0.88$}& 37.46 \tiny{$\pm 0.84$}
                     
                     & 26.52 \tiny{$\pm 0.31$}& 25.71 \tiny{$\pm 0.44$}
                  
                     \\
                     
                     & DR-GCN 
                     & 34.84 \tiny{$\pm 0.72$}& 31.53 \tiny{$\pm 0.86$}
                    
                     & 24.69 \tiny{$\pm 0.46$}& 21.81 \tiny{$\pm 0.42$}
                    
                     \\

                     & GraphENS
                     & 39.71 \tiny{$\pm 0.55$}& 38.75 \tiny{$\pm 0.60$}
                    
                     & 26.55 \tiny{$\pm 0.49$}& 26.00 \tiny{$\pm 0.52$}
                   
                     \\
                     
                     \cline{2-6}
                     
                     & BalancedSoftmax 
                     & 39.27 \tiny{$\pm 0.83$}& 38.53 \tiny{$\pm 0.87$}
                    
                     & 26.09 \tiny{$\pm 0.43$}& 25.28 \tiny{$\pm 0.38$}
                    
                     \\
                     & + \textbf{TAM} 
                     & \textbf{41.40} \tiny{$\pm 0.57$}& 40.25 \tiny{$\pm 0.72$}
                    
                     & 26.91 \tiny{$\pm 0.36$}& 26.19 \tiny{$\pm 0.38$}
                  
                     \\
                     \cdashline{2-6}
                     & ReNode
                     & 37.95 \tiny{$\pm 0.78$}& 37.09 \tiny{$\pm 0.87$}
                     
                     & 26.14 \tiny{$\pm 0.52$}& 25.47 \tiny{$\pm 0.52$}
                     \\
                     
                     & + \textbf{TAM} 
                     & 37.57 \tiny{$\pm 0.97$}& 36.11 \tiny{$\pm 0.96$}
                     
                     & 26.08 \tiny{$\pm 0.41$}& 25.39 \tiny{$\pm 0.37$}
                     
                     \\
                     \cdashline{2-6}
                     & GraphSMOTE 
                     & 40.18 \tiny{$\pm 0.67$}& 39.43 \tiny{$\pm 0.76$}
                    
                     & \textbf{27.10} \tiny{$\pm 0.49$}& \textbf{26.63} \tiny{$\pm 0.63$}
                   
                     \\
                     & + \textbf{TAM} 
                     & 41.19 \tiny{$\pm 0.55$}& \textbf{40.51} \tiny{$\pm 0.68$}
                    
                     & 26.56 \tiny{$\pm 0.46$} & 25.74 \tiny{$\pm 0.47$}
                    
                     \\
\cline{2-6}
\noalign{\vskip\doublerulesep
         \vskip-\arrayrulewidth} \cline{2-6}
\rule{0pt}{2.5ex}  
\multirow{11}{*}{\rotatebox{90}{SAGE}} & Cross Entropy 
                    & 32.07 \tiny{$\pm 0.48$}& 25.33 \tiny{$\pm 0.73$}
                  
                    & 25.55 \tiny{$\pm 0.41$}& 20.29 \tiny{$\pm 0.41$}
                   
                    \\
                    \rule{0pt}{2ex}
                     & Re-Weight
                     & 36.49 \tiny{$\pm 1.21$}& 34.84 \tiny{$\pm 1.30$}
                     
                     & 29.83 \tiny{$\pm 0.59$}& 25.88 \tiny{$\pm 0.42$}
                    
                     \\
                    
                     & PC Softmax 
                     & 40.71 \tiny{$\pm 0.82$}& 39.95 \tiny{$\pm 0.98$}
                     
                     & 29.23 \tiny{$\pm 0.50$}& 28.19 \tiny{$\pm 0.54$}
                    
                     \\
                     
                     & DR-GCN 
                     & 37.24 \tiny{$\pm 0.79$}& 34.37 \tiny{$\pm 0.97$}
                   
                     & 28.77 \tiny{$\pm 0.70$}& 22.32 \tiny{$\pm 0.96$}
                    
                     \\

                     & GraphENS 
                     & 34.91 \tiny{$\pm 0.68$}& 33.47 \tiny{$\pm 0.81$}
                     
                     & 24.09 \tiny{$\pm 0.36$}& 23.03 \tiny{$\pm 0.32$}
                    
                     \\
                     
                     \cline{2-6}
                     
                     & BalancedSoftmax 
                     & 40.76 \tiny{$\pm 0.99$}& 40.27 \tiny{$\pm 1.05$}
                     
                     & \textbf{30.07} \tiny{$\pm 0.44$}& \textbf{29.12} \tiny{$\pm 0.42$}
                    
                     \\
                     & + \textbf{TAM} 
                     & \textbf{41.19} \tiny{$\pm 1.08$}& \textbf{40.41} \tiny{$\pm 1.13$}
                     
                     & 29.91 \tiny{$\pm 0.50$}& 28.56 \tiny{$\pm 0.58$}
                    
                     \\
                     \cdashline{2-6}
                     & ReNode 
                     & 37.07 \tiny{$\pm 1.02$}& 36.02 \tiny{$\pm 0.99$}
                    
                     & 29.48 \tiny{$\pm 0.64$}& 26.09 \tiny{$\pm 0.58$}
                     
                     \\
                     & + \textbf{TAM}
                     & 38.24 \tiny{$\pm 0.93$}& 37.23 \tiny{$\pm 1.09$}
                    
                     & 29.77 \tiny{$\pm 0.58$} & 27.72 \tiny{$\pm 0.68$}
                    
                     \\
                     \cdashline{2-6}
                     & GraphSMOTE 
                     & 33.31 \tiny{$\pm 0.63$}& 30.83 \tiny{$\pm 0.67$}
                     
                     & 25.51 \tiny{$\pm 0.43$}& 19.79 \tiny{$\pm 0.49$}
                    
                     \\
                     & + \textbf{TAM}
                     & 33.23 \tiny{$\pm 0.54$}& 30.66 \tiny{$\pm 0.74$}
                     
                     & 25.34 \tiny{$\pm 0.56$} & 22.29 \tiny{$\pm 0.48$}
                     
                     \\

\bottomrule

\end{tabular}
\end{adjustbox}
\end{footnotesize}
\end{center}
\label{tb:appx_main_hetero}
\end{table*}

\clearpage

\section{Detailed Experimental Results} \label{appen:experiments}
In this section, we describe detailed experimental settings: dataset statistics, evaluation protocol, and implementation details.

\subsection{Label Distribution in Training Datasets}
We provide the label distribution in class-imbalanced datasets in Table~\ref{tb:datastat}.

\begin{table}[h] 
\center
\caption{\small Label distribution in training datasets [\%]}
\vspace{-0 in}
\begin{small}
\setlength{\tabcolsep}{3pt} 
\begin{tabular}{l|ccccccc}
\toprule
    \textbf{Dataset} & $\mathbf{L}_0$ & $\mathbf{L}_1$ & $\mathbf{L}_2$ & $\mathbf{L}_3$ & $\mathbf{L}_4$ & $\mathbf{L}_5$ & $\mathbf{L}_6$ \\
    \hline
    Cora ($\rho=5$) & 21.74 & 21.74 & 21.74 & 21.74 & 4.35 & 4.35 & 4.35 \\
    Cora ($\rho=10$) & 23.26 & 23.26 & 23.26 & 23.26 & 2.33 & 2.33 & 2.33 \\
    \cline{0-7}
    CiteSeer ($\rho=5$) & 27.78 & 27.78 & 27.78 & 5.56 & 5.56 & 5.56 &-  \\
    CiteSeer ($\rho=10$) & 30.30 & 30.30 & 30.30 & 3.03 & 3.03 & 3.03 &-  \\
    \cline{0-7}
    PubMed ($\rho=5$) & 45.45 & 45.45 & 9.09 &- &- &- &- \\
    PubMed ($\rho=10$) & 47.62 & 47.62 & 4.76 &- &- &- &- \\
    \cline{0-7}
    Chameleon ($\rho=5$) & 29.51 & 29.31 & 29.03 & 6.07 & 6.07 & - & - \\
    Chameleon ($\rho=10$) & 31.44 & 31.22 & 30.93 & 3.20 & 3.20 & - & - \\
    \cline{0-7}
    Squirrel ($\rho=5$) & 29.54 & 29.53 & 29.07 & 5.98 & 5.98 & - & -  \\ 
    Squirrel ($\rho=10$) & 31.43 & 31.31 & 30.93 & 3.17 & 3.17 & - & -  \\ 
    \cline{0-7}
    Wisconsin ($\rho=11.63$) & 46.50 & 27.92 & 13.42 & 8.17 & 4.00 & - & -  \\ 
  
\bottomrule
\end{tabular}
\end{small}
\label{tb:datastat}
\vspace{-0.1in}
\end{table}

\subsection{Evaluation Protocol} 
We validate our algorithm and baselines on various GNN architectures: GCN~\citep{gcn}, GAT~\citep{gat} and GraphSAGE~\citep{sage}. We follow the detailed architecture used in \citet{renode}. All GNNs consist of their own convolutional layers with ReLU activation and dropout~\citep{dropout} is applied with dropping rate of 0.5 before the last layer. For 1-layer GNNs, we do not adopt dropout and we use multi-head attention with 4 heads for GAT. We search the best architecture based on the average of validation accuracy and F1 score among the number of layers $l\in\{1,2,3\}$ and the hidden dimension $d\in\{64,128,256\}$. For optimization, we train models for 2000 epochs with Adam optimizer~\citep{adam}. The initial learning rate is set to 0.01 and the learning rate is halved if the validation loss has not improved for 100 iterations. Weight decay is applied to all learnable parameters as 0.0005 except for the last convolutional layer.

\subsection{Implementation Details} 
For our algorithm, we search the best hyperparameters based on the average of validation accuracy and F1 among the coefficient of ACM term $\alpha\in\{0.25,0.5,1.5,2.5\}$, the coefficient of ADM term $\beta\in\{0.125,0.25,0.5\}$, and the minimum temperature of class-wise temperature $\phi\in\{0.8,1.2\}$. The sensitivity to imbalance ratio of class-wise temperature $\delta$ is fixed as 0.4 for all main experiments. We adopt warmup for 5 iterations since we utilize model prediction for unlabeled nodes.

\subsection{Baselines} \label{appensub:baselines}
For DR-GCN~\citep{conditionalgan}, we only utilize the module to keep representations having structure information with conditional GAN and do not adopt the component, which exploits unlabeled nodes, mainly targeting semi-supervised learning for a fair comparison. For GraphSMOTE~\citep{graphsmote}, we select the model whose predicted edges have discrete values among multiple versions in that this setting shows superior performance in many datasets. Since the interpolation happens in the representation space, we search the best architecture for GraphSMTOE among the number of layers $l\in\{2,3\}$. For ReNode~\citep{renode}, we search hyperparameters among lower bound of cosine annealing $w_{min}\in\{0.25,0.5,0.75\}$ and upper bound of the cosine annealing $w_{max}\in\{1.25,1.5,1.75\}$ as following \citet{renode}. PageRank teleport probability is fixed as $\alpha=0.15$, which is the default setting in the released codes.


\end{document}